\RequirePackage[T1]{fontenc}
\documentclass[10pt,journal,twoside,letterpaper]{IEEEtran}

\ifdefined\IEEEreviewcopy
  \usepackage[letterpaper,margin=1in,bottom=1.12in,includehead]{geometry}
\fi
\usepackage{amsmath,amssymb,booktabs,array,graphicx,cite,url}
\usepackage{algorithm}
\usepackage{algorithmic}
\newtheorem{assumption}{\textbf{Assumption}}
\newtheorem{lemma}{\textbf{Lemma}}
\newtheorem{theorem}{\textbf{Theorem}}
\newtheorem{corollary}{\textbf{Corollary}}
\newtheorem{remark}{\textbf{Remark}}
\newcommand{\bestresult}[1]{\boldsymbol{#1}}
\newcommand{\E}{\mathbb E}
\newcommand{\R}{\mathbb R}
\newcommand{\gh}{\widehat g}
\newcommand{\sg}{\widetilde\nabla}
\newcommand{\prox}{\operatorname{prox}}
\newcommand{\vecop}{\operatorname{vec}}
\newcommand{\Tset}{\mathcal T}
\newcommand{\Rset}{\mathcal R}
\newcommand{\Mset}{\mathcal M}
\newcommand{\paperTitle}{Projected Penalty-Gradient Tracking for Communication-Efficient Byzantine-Resilient Federated Learning}
\newcommand{\printreferences}{\bibliographystyle{IEEEtran}\bibliography{references_polished}}

\begin{document}
\title{\paperTitle}
\author{Yanjie Dong%
}
\maketitle
\begin{abstract}
We investigate communication-efficient federated learning over heterogeneous datasets in the presence of Byzantine adversaries capable of uploading arbitrary information.
Penalty-based methods bound the influence of each worker through the bounded gradient of a penalty function; however, the bound is lost once the uploads are compressed, since a Byzantine worker can accumulate an unbounded state at the server.
To retain the Byzantine resilience under compressed uploads, we propose a projected penalty-gradient tracking framework, where each honest worker uploads a compressed residual of its penalty gradient and the server projects each reconstructed state onto the known penalty-gradient ball.
The framework accommodates any contractive compressor and yields two algorithms: fault-resilient proximal gradient with consistent sparsification (FRPG-CS), whose shared mask is protected by freshly drawn reserve rows, and lazily aggregated FRPG (LAG-FRPG), whose uploads are event-triggered.
For smooth and strongly convex losses, we establish a finite-time bound on the expected objective gap, where the Byzantine error floor is independent of the compressor and the compression error decays faster than the optimization error.
We also bound the expected number of lazy uploads and derive certificates that separate the finite-penalty model bias from the optimization error.
Experiments on MNIST and Fashion-MNIST show that the proposed algorithms attain the highest accuracy among twelve methods under the Min-Max attack, whereas a compressed-momentum benchmark and a local-update benchmark remain preferable in attack-free accuracy and total traffic, respectively.
\end{abstract}

\section{Introduction}
Federated learning (FL) enables a group of workers to learn from decentralized data under the orchestration of a central server, where each worker performs local computation and exchanges only model-related information with the server~\cite{mcmahan2017}.
When FL is deployed over a large number of heterogeneous devices, it faces two major challenges, namely, Byzantine attacks~\cite{blanchard2017,yin2018} and the communication bottleneck~\cite{stich2018,chen2018lag}.
Byzantine adversaries may upload arbitrary information to disrupt the learning process, while the frequent upload of high-dimensional vectors over bandwidth-limited uplinks can make the training prohibitively slow.

Current Byzantine-resilient FL algorithms either aggregate the uploaded gradients by robust rules~\cite{blanchard2017,yin2018,mhamdi2018bulyan,pillutla2022rfa,wu2020byrdsaga} or penalize the disagreement between the local model of each worker and the server model~\cite{li2019rsa,dong2020frpg}.
The penalty-based methods, e.g., robust stochastic aggregation (RSA)~\cite{li2019rsa} and fault-resilient proximal gradient (FRPG)~\cite{dong2020frpg}, are naturally suited to heterogeneous datasets, since the influence of each worker on the server model is bounded by the gradient of the penalty function.

Combining the penalty-based methods with communication compression is nontrivial.
The bound on the penalty gradient holds for the vector that an honest worker computes rather than for the vector that a Byzantine worker uploads.
Moreover, communication-efficient algorithms accumulate compressed increments into a persistent state at the server~\cite{stich2018,karimireddy2019ef,richtarik2021}; projecting each received increment is thus insufficient, since the accumulated state of a Byzantine worker can still grow.
Therefore, we track the penalty gradients with projected states: each honest worker uploads a compressed residual between its exact penalty gradient and a local copy of its server-side state, while the server projects each accumulated state onto the known penalty-gradient ball.
In this way, the state that enters the server update remains bounded regardless of the uploaded payloads.

We focus on smooth and strongly convex losses, which cover $\ell_2$-regularized generalized linear models and least-squares estimators, e.g., the logistic regression in a production federated keyboard~\cite{yang2018gboard}, linear heads on frozen pretrained models~\cite{legate2023lastlayer}, parameter estimation from heterogeneous and partly compromised measurements~\cite{chen2019sage}, and the state estimation of power grids under false data injection~\cite{liu2011fdi}.
Strong convexity also makes the learning target unique, such that an error floor in the objective gap bounds how far the Byzantine workers can move the model.

\subsection{Related Works}
\paragraph{Byzantine resilience}
Since the seminal work~\cite{blanchard2017}, the standard mean has been replaced by robust aggregation rules, e.g., Krum~\cite{blanchard2017}, coordinate-wise median and trimmed mean~\cite{yin2018}, Bulyan~\cite{mhamdi2018bulyan}, and geometric median~\cite{pillutla2022rfa}.
Their original guarantees rely on concentration, common-distribution, or honest-majority conditions, and cannot be transferred unchanged to heterogeneous datasets.
To handle the data heterogeneity, worker momentum with centered clipping~\cite{karimireddy2021history}, bucketing~\cite{karimireddy2022bucketing}, and nearest-neighbor mixing~\cite{allouah2023mixing} have been developed, while the tight bounds in~\cite{allouah2023breakdown,shi2025optimal} show that the achievable learning error depends on the heterogeneity model.
Other works rely on trusted data or reference gradients at the server~\cite{cao2021fltrust,xie2020zenopp,shestakov2026jl}.
Meanwhile, the inner-product manipulation (IPM)~\cite{xie2019ipm}, a-little-is-enough (ALIE)~\cite{baruch2019alie}, and Min-Max~\cite{shejwalkar2021minmax} attacks show that the knowledge and the feasible payloads of the adversaries need to be specified.

Different from the above methods that minimize the consensus loss of the honest workers, the penalty-based methods allow the local models to differ from the server model (see also~\cite{koppel2017,dinh2020pfedme}).
RSA~\cite{li2019rsa} employs the bounded subgradients of a norm penalty and separates the consensus error from the penalty bias, while FRPG and its local variant LFRPG~\cite{dong2020frpg} accelerate the penalty-based route via proximal-gradient recursions.
Nevertheless, RSA and FRPG upload a dense vector per worker and slot, and the guarantee of FRPG requires the messages of the Byzantine workers to obey the norm bound of the penalty gradient.

\paragraph{Communication compression}
Sparsification with memory~\cite{stich2018} and error feedback~\cite{karimireddy2019ef} correct the bias of contractive compressors, where the residual bound in~\cite[Lemma~3]{karimireddy2019ef} scales as $(1-\gamma)/\gamma^2$ with the contraction factor $\gamma$ and vanishes at full communication.
EF21~\cite{richtarik2021} tracks a changing vector by compressed differences and provides the tracking recurrence that we build on.
In the presence of Byzantine adversaries, compression has been combined with robust aggregation via error feedback~\cite{ghosh2021ef}, gradient-difference compression in BROADCAST~\cite{zhu2023broadcast}, Byz-DASHA-PAGE and Byz-EF21~\cite{rammal2024}, compressed momentum filtering~\cite{liu2026momentum}, Byz-DM21~\cite{li2026byzdm21}, coordinated sparse supports in RoSDHB~\cite{gupta2026rosdhb}, and globally consistent Top-$K$ compression in FedAdam-BACK~\cite{xu2026fedadamback}; a common support is also selected by ARC-Top-$K$~\cite{chen2025arctopk} for All-Reduce compatibility.
For the penalty-based methods, C-RSA~\cite{he2023crsa} combines compression with RSA in the non-convex and non-i.i.d. regime.
Nevertheless, the tracking of penalty gradients with bounded server-side states has yet to be characterized for the accelerated penalty-based methods.

\paragraph{Lazy and event-triggered communication}
LAG~\cite{chen2018lag} reuses the outdated gradients of the workers whose gradients change slowly, LASG~\cite{chen2021lasg} adapts the triggering rules to stochastic gradients, and LAQ~\cite{sun2022laq} combines lazy uploads with quantization.
Event triggering is combined with compression in SPARQ-SGD~\cite{singh2023sparq} and analyzed asymptotically for FL in~\cite{he2023event}, and the communication frequency is also reduced by local updates in ProxSkip~\cite{mishchenko2022proxskip}, LoCoDL~\cite{condat2025locodl}, and LFRPG~\cite{dong2020frpg}.
Nevertheless, to the best of our knowledge, lazy uploads have not been analyzed in the presence of Byzantine adversaries.

\subsection{Our Contributions}
Motivated by the above limitations, we develop a communication-efficient penalty-based FL framework whose Byzantine resilience does not degrade with the communication format.
Our contributions are as follows.
\begin{itemize}
    \item We propose a projected penalty-gradient tracking framework, where each honest worker uploads a compressed residual of its penalty gradient and the server projects each reconstructed state onto the penalty-gradient ball.
    The framework accommodates any contractive compressor and bounds the state of each Byzantine worker regardless of the uploaded payloads.
    In particular, FRPG-CS lets all workers share a mask that combines robustly ranked rows with freshly drawn reserve rows, such that the contraction is guaranteed even when the ranking is manipulated, and LAG-FRPG lets each honest worker upload only when its residual exceeds a decaying threshold.
    \item For smooth and strongly convex losses, we establish a finite-time bound on the expected objective gap without assuming bounded stochastic gradients or an honest majority.
    The Byzantine error floor is independent of the compressor, while the compression adds an ${\cal O}(\ln K/K^2)$ term over $K$ slots, whose dependence $(1-\gamma)/\gamma^2$ on the contraction factor cannot be improved in order.
    For LAG-FRPG, the lazy uploads add an ${\cal O}(1/(K\ln K))$ term, and the expected fraction of uploading slots decays as ${\cal O}(\sqrt{\ln K/K})$.
    \item We derive computable certificates that separate the finite-penalty model bias from the optimization error for any penalty weight, and evaluate the proposed algorithms against ten benchmarks on the MNIST and Fashion-MNIST datasets.
\end{itemize}

\textbf{Organization.}~The remaining work is organized as follows.
The problem description and several preliminaries are introduced in Section~\ref{sec:problem}.
The algorithmic development and the theoretical results are provided in Sections~\ref{sec:algorithm} and~\ref{sec:convergence}, respectively.
Numerical results are presented in Section~\ref{sec:tests}, which is followed by the concluding remarks in Section~\ref{sec:conclusion}.

\section{Problem Description and Preliminaries}\label{sec:problem}
\subsection{Penalty-Based Federated Learning}
We consider an FL system where a central server orchestrates $Q=N+B$ workers, among which $N$ workers are honest and $B$ workers are Byzantine.
Without loss of generality, we index the honest workers by $n=1,\ldots,N$; the indexing is only used in the analysis, and the server requires neither the identities of the honest workers nor the value of $B$.
Each honest worker $n$ holds a local dataset $\mathcal D_n$ with the local loss $f_n(x):=\E_{\xi\sim\mathcal D_n}[f(x;\xi)]$, and the server holds a regularizer $f_0$.
With equally weighted workers, the consensus-based FL minimizes
\begin{equation}\label{eq:consensus}
x^\star=\arg\min_{x\in\R^d}H(x):=f_0(x)+\sum_{n=1}^Nf_n(x)
\end{equation}
where $x$ is the $d$-dimensional model parameter vector.
To handle the heterogeneous datasets and bound the influence of each worker, the penalty-based methods~\cite{li2019rsa,dong2020frpg} assign a local model $w_n\in\R^d$ to each honest worker $n$ and a model $w_0\in\R^d$ to the server, and minimize
\begin{equation}\label{eq:penalized}
 F(w):=f_0(w_0)+\sum_{n=1}^N\{f_n(w_n)+\lambda p_n(w_0-w_n)\}
\end{equation}
where $w:=\vecop(w_0,\ldots,w_N)$, $\lambda>0$ is the penalty weight, and $p_n$ is the penalty function of worker $n$.
We denote the minimizer of~\eqref{eq:penalized} by $u^\ast:=\vecop(u_0^\ast,\ldots,u_N^\ast)$, and rely on the following assumptions.

\begin{assumption}[Smoothness]\label{as:lipschitz}
For $n=0,\ldots,N$, the function $f_n$ has an $L_n$-Lipschitz continuous gradient.
\end{assumption}
\begin{assumption}[Strong convexity]\label{as:strong}
For $n=0,\ldots,N$, the function $f_n$ is $\delta_n$-strongly convex with $\delta_n>0$.
\end{assumption}
\begin{assumption}[Bounded penalty gradient]\label{as:penalty}
For $n=1,\ldots,N$, the penalty function $p_n$ is convex and differentiable with $\|\nabla p_n(y)\|^2\le G$ and an $L_p$-Lipschitz continuous gradient, where $G>0$.
\end{assumption}

Assumptions~\ref{as:lipschitz} and~\ref{as:strong} hold for $\ell_2$-regularized generalized linear models and least-squares estimators, where a worker with a rank-deficient local design requires a positive ridge regularization since Assumption~\ref{as:strong} is imposed per worker.
Assumption~\ref{as:penalty} is satisfied by the radial Huber penalty
\begin{equation}\label{eq:huber}
p(y)=\begin{cases}
	\|y\|^2/(2\mu)	&\|y\|\le\mu \\
	\|y\|-\mu/2,	&\|y\|>\mu\end{cases}
\end{equation}
with $\mu>0$, $G=1$, and $L_p=1/\mu$, which we use for all workers in the sequel.
Assumption~\ref{as:penalty} implies that each honest worker moves the server model through the penalty gradient $\lambda\nabla p_n(w_0-w_n)$, whose norm never exceeds $\lambda\sqrt G$.

\subsection{Threat Model}
The Byzantine workers can collude, know the algorithm, observe the messages and the current vectors of the honest workers, and upload arbitrary finite real-valued vectors, where a missing upload is replaced by zero.
The server is trusted, i.e., it follows the protocol, accepts only the coordinates permitted by the announced communication format, and rejects malformed messages.
The identities of the workers are fixed, and the links between the honest workers and the server are synchronous and reliable, such that each honest worker and the server keep identical copies of the state of that worker; packet loss without state reconciliation is beyond the scope of this work.
Moreover, the randomness drawn by the server is independent of the history, i.e., the Byzantine workers can react to an announced mask but cannot predict it when committing their earlier reports.

\subsection{Finite-Penalty Target and Consensus Certificates}
For a finite penalty weight $\lambda$, the server optimum $u_0^\ast$ of~\eqref{eq:penalized} generally differs from the consensus optimum $x^\star$ in~\eqref{eq:consensus}.
In fact, $u_0^\ast$ minimizes the profiled loss $f_0(x)+\sum_{n=1}^N\inf_y\{f_n(y)+\lambda p(x-y)\}$, to which each worker contributes a gradient with norm at most $\lambda$ (cf. Appendix~\ref{app:consensus}); hence, the bounded influence of each worker comes with a model bias.
Let $\Pi_r$ denote the Euclidean projection onto the ball with radius $r$, $\delta_\Sigma:=\sum_{n=0}^N\delta_n$, and $\delta_{\min}:=\min_n\delta_n$.
The model bias is quantified in the following lemma, whose proof is deferred to Appendix~\ref{app:consensus}.

\begin{lemma}[Unconditional consensus certificates]\label{lem:consensus}
Suppose Assumptions~\ref{as:lipschitz}--\ref{as:penalty} are satisfied.
For any $w=\vecop(w_0,\ldots,w_N)$, let $b(w):=\|\sum_{n=0}^N\nabla_nF(w)\|+\sum_{n=1}^NL_n\|w_0-w_n\|$ with $\nabla_nF$ being the gradient of $F$ with respect to $w_n$.
Then, we have
\begin{equation}\label{eq:consensus-residual-distance}
\|w_0-x^\star\|\le\frac{\|\nabla H(w_0)\|}{\delta_\Sigma}
\le\frac{b(w)}{\delta_\Sigma}
\end{equation}
and
\begin{equation}\label{eq:consensus-residual-objective}
H(w_0)-H(x^\star)\le\frac{\|\nabla H(w_0)\|^2}{2\delta_\Sigma}.
\end{equation}

Moreover, let $e_n:=\nabla f_n(x^\star)-\Pi_\lambda(\nabla f_n(x^\star))$ and $A_\lambda^2:=\|\sum_{n=1}^Ne_n\|^2/\delta_0+\sum_{n=1}^N\|e_n\|^2/\delta_n$.
For the Huber penalty~\eqref{eq:huber} with any $\lambda,\mu>0$, the server optimum satisfies
\begin{equation}\label{eq:unconditional-huber-envelope}
\|u_0^\ast-x^\star\|\le
(A_\lambda+\sqrt{N\lambda\mu})/\sqrt{\delta_0}.
\end{equation}
\end{lemma}

The inequalities~\eqref{eq:consensus-residual-distance} and~\eqref{eq:consensus-residual-objective} certify the distance and the objective gap between any server model and the consensus optimum; evaluated at an approximate minimizer $\bar u$ of $F$, the certificate~\eqref{eq:consensus-residual-distance} holds for $u_0^\ast$ after adding $\|\nabla F(\bar u)\|/\delta_{\min}$ to its right-hand side (RHS).
Since the certificates use the full-data quantities of the honest workers, they serve as offline diagnostics rather than information available to the server.

\begin{remark}[Penalty weight and model bias]\label{rem:bias}
When $\lambda>\max_n\|\nabla f_n(u_n^\ast)\|$, the penalized optimum lies in the quadratic region of the Huber penalty, and the sharper bound $\|u_0^\ast-x^\star\|\le\mu\sum_{n=1}^NL_n/\delta_\Sigma$ holds (cf. Appendix~\ref{app:consensus}).
However, such a condition fails in all ten configurations with the Dirichlet concentration $0.3$ in Section~\ref{sec:bias}, whereas~\eqref{eq:unconditional-huber-envelope} holds for any $\lambda$ and $\mu$.
If $\lambda\ge\max_n\|\nabla f_n(x^\star)\|$, we have $A_\lambda=0$ and the model bias vanishes with $\mu$; otherwise, reducing $\mu$ alone cannot remove the model bias in general.
Meanwhile, a larger $\lambda$ raises the Byzantine error floor (cf. Theorem~\ref{thm:cs}), which reveals a tradeoff between the model bias and the Byzantine resilience.
\end{remark}

\section{Projected Penalty-Gradient Tracking}\label{sec:algorithm}
In this section, we develop the projected penalty-gradient tracking framework based on the FRPG recursions~\cite{dong2020frpg}, and specify the communication formats of the FRPG-CS and LAG-FRPG algorithms.

\subsection{Projected Tracking Framework}
Per slot $k\ge1$, the server and the honest workers perform the FRPG recursions with the stepsizes
\begin{equation}\label{eq:beta}
\beta_k=2/(k+2)
\end{equation}
and
\begin{equation}\label{eq:alpha}
\alpha_{n,k}=\begin{cases}\delta_0(k+2)^2/14+3L_0/2,&n=0,\\3\delta_n(k+2)^2/14+L_n,&n\ge1.\end{cases}
\end{equation}

More specifically, the server updates
\begin{subequations}\label{eq:server}
\begin{align}
u_{0,k}&=(1-\beta_k)w_{0,k-1}+\beta_k v_{0,k-1}\label{eq:server-u}\\
w_{0,k}&=u_{0,k}-\nabla f_0(u_{0,k})/\alpha_{0,k}\label{eq:server-w}\\
v_{0,k}&=v_{0,k-1}-\frac{\delta_0(v_{0,k-1}-u_{0,k})+\nabla f_0(u_{0,k})+S_k}{\delta_0+\alpha_{0,k}\beta_k}\label{eq:server-v}
\end{align}
\end{subequations}
where $S_k:=\sum_{n=1}^Q\gh_{n,k}$ aggregates the states $\{\gh_{n,k}\}_{n=1}^Q$ that the server maintains for all workers.
After receiving $w_{0,k}$, each honest worker $n$ samples a mini-batch $\xi_{n,k}$ from $\mathcal D_n$ and updates
\begin{subequations}\label{eq:worker}
\begin{align}
u_{n,k}&=(1-\beta_k)w_{n,k-1}+\beta_k v_{n,k-1} \label{eq:worker-u}\\
w_{n,k}&=w_{0,k}-\prox_{\frac{\lambda p_n}{\alpha_{n,k}}}\bigl( w_{0,k}-u_{n,k}+\sg_{n,k}/\alpha_{n,k} \bigr) \label{eq:worker-w}\\
v_{n,k}&=v_{n,k-1}-\frac{\delta_n(v_{n,k-1}-u_{n,k})+\sg_{n,k}-g_{n,k}}{\delta_n+\alpha_{n,k}\beta_k} \label{eq:worker-v}
\end{align}
\end{subequations}
where $\sg_{n,k}:=\nabla f(u_{n,k};\xi_{n,k})$ is the stochastic gradient, $\prox_{tp}(x):=\arg\min_y\{tp(y)+\|x-y\|^2/2\}$ is the proximal operator, and
\begin{equation}\label{eq:penaltygrad}
g_{n,k}:=\lambda\nabla p_n(w_{0,k}-w_{n,k}),\qquad
z_{n,k}:=g_{n,k}-\gh_{n,k-1}
\end{equation}
are the penalty gradient of worker $n$ and its residual with respect to the state $\gh_{n,k-1}$ that the server maintains for worker $n$.
Note that~\eqref{eq:worker-v} uses the exact penalty gradient $g_{n,k}$, which is locally available regardless of the uploaded information.

Different from FRPG, where each worker uploads $g_{n,k}$ per slot, each honest worker in the proposed framework uploads a compressed version $c_{n,k}$ of the residual $z_{n,k}$ and keeps an identical copy of its state.
After receiving $c_{n,k}$, the server updates the state of each worker $n=1,\ldots,Q$ as
\begin{equation}\label{eq:state}
\gh_{n,k}=\Pi_{\lambda\sqrt G}(\gh_{n,k-1}+c_{n,k})
\end{equation}
with $\gh_{n,0}=0$, while the local models are not projected.
The projection in~\eqref{eq:state} plays two roles.
For an honest worker, it cannot increase the tracking error $\|\gh_{n,k}-g_{n,k}\|$, since $g_{n,k}$ belongs to the ball with radius $\lambda\sqrt G$ under Assumption~\ref{as:penalty}.
For a Byzantine worker, the state obeys the same norm bound regardless of the uploaded payloads.
The detailed procedures are summarized in Algorithm~\ref{alg:tracking}, where the models $w_{n,0}$ and $v_{n,0}$ are initialized at deterministic finite vectors.

\begin{algorithm}[!t]
\caption{Projected Penalty-Gradient Tracking}
\label{alg:tracking}
\begin{algorithmic}[1]
\REQUIRE stepsizes $\{\alpha_{n,k}\}$ and $\{\beta_k\}$; penalty weight $\lambda$; communication format; number of slots $K$.
\STATE Server and honest workers initialize $w_{n,0}$, $v_{n,0}$, and $\gh_{n,0}=0$
\FOR{$k=1,\ldots,K$}
\STATE Server computes $u_{0,k}$ and $w_{0,k}$ via~\eqref{eq:server-u} and~\eqref{eq:server-w}, and broadcasts $w_{0,k}$ to all workers
\STATE Each honest worker $n$ samples $\xi_{n,k}$ and computes $u_{n,k}$, $w_{n,k}$, $v_{n,k}$, and $z_{n,k}$ via~\eqref{eq:worker} and~\eqref{eq:penaltygrad}
\IF{FRPG-CS}
\STATE Each worker uploads its row energies; server clips them via~\eqref{eq:clip}, selects $\Tset_k$ by trimmed means, draws the reserve $\Rset_k$, and broadcasts $\Mset_k$
\STATE Each honest worker uploads $c_{n,k}$ via~\eqref{eq:mask}
\ELSIF{LAG-FRPG}
\STATE Each honest worker uploads $c_{n,k}$ via~\eqref{eq:lazy}
\ELSE
\STATE Each honest worker uploads $c_{n,k}={\cal C}_{n,k}(z_{n,k})$ and the required support information
\ENDIF
\STATE Server updates the states of all workers via~\eqref{eq:state}, and each honest worker applies the same update to its copy
\STATE Server computes $v_{0,k}$ via~\eqref{eq:server-v}
\ENDFOR
\RETURN $w_{0,K}$
\end{algorithmic}
\end{algorithm}

\subsection{Contractive Compressors}
Let $c_{n,k}={\cal C}_{n,k}(z_{n,k})$ with ${\cal C}_{n,k}$ being the compressor of worker $n$ per slot $k$, and let ${\cal H}_k$ denote the history that fixes the residuals of the honest workers and precedes the fresh randomness of the compressors.
The proposed framework requires each honest worker $n$ to use a contractive compressor, i.e.,
\begin{equation}\label{eq:general-contraction}
\E[\|\mathcal C_{n,k}(z_{n,k})-z_{n,k}\|^2\mid\mathcal H_k]
\le(1-\gamma_n)\|z_{n,k}\|^2
\end{equation}
where $\gamma_n\in(0,1]$ is the contraction factor.
The compressor may vary with the history, and no independence among the workers is required.

To specify the compressors, we zero-pad and partition each $d$-dimensional vector into $m:=\lceil d/\ell\rceil$ rows of length $\ell$.
For an integer $1\le r\le m$, keeping the $r$ rows with the largest energy of each residual (individual top-$r$ rows) and keeping $r$ uniformly sampled rows that are shared by all workers both satisfy~\eqref{eq:general-contraction} with $\gamma_n=r/m$.
The former requires each worker to upload the indices of its rows, the latter requires the server to broadcast a mask, and the full upload corresponds to $\gamma_n=1$.
Note that each honest worker mirrors the entire projected state in~\eqref{eq:state}, since the projection may rescale the coordinates outside the transmitted support.

\subsection{Consistent Sparsification (FRPG-CS)}
The FRPG-CS algorithm lets all workers share a data-dependent mask, which combines robustly ranked rows with freshly drawn reserve rows since the ranking can be manipulated by the Byzantine workers.
Based on the row partition, each honest worker first uploads the row energies $\psi_{n,k}[j]:=\|z_{n,k}[j]\|^2$ with $z_{n,k}[j]$ being the $j$-th row of $z_{n,k}$, $j=1,\ldots,m$.
Since the residual of an honest worker satisfies $\|z_{n,k}\|^2\le4\lambda^2G$, the server clips each report as
\begin{equation}\label{eq:clip}
\widetilde\psi_{n,k}=
\frac{4\lambda^2G[\psi_{n,k}]_+}{\max\{4\lambda^2G,\sum_j[\psi_{n,k}[j]]_+\}}
\end{equation}
where $[\cdot]_+$ denotes the entry-wise positive part.
For each row, the server computes a trimmed mean of the clipped reports with $\widehat b$ entries removed from each tail, where $0\le\widehat b<Q/2$.
The $s$ rows with the highest scores form the set $\Tset_k$ with index-based tie breaking.
After the reports are fixed, the server selects $R\ge1$ rows uniformly without replacement from the complement of $\Tset_k$ to form the reserve set $\Rset_k$.
The mask and the upload of each honest worker are
\begin{equation}\label{eq:mask}
\Mset_k=\Tset_k\cup\Rset_k,\quad s+R\le m,\quad
c_{n,k}=P_{\Mset_k}z_{n,k}
\end{equation}
where $P_{\Mset_k}$ keeps the rows in $\Mset_k$ and sets the remaining rows to zero.
Hence, each honest worker uploads $m+(s+R)\ell$ scalars per slot, and the mask is broadcast on the downlink.

Let $\mathcal F'_k$ denote the history that includes the current residuals of the honest workers and all energy reports, but excludes the fresh reserve and the subsequent payloads.
Since $\Tset_k$ and $z_{n,k}$ are fixed given $\mathcal F'_k$, we have
\begin{equation}\label{eq:contraction}
\E[\|z_{n,k}-c_{n,k}\|^2\mid\mathcal F'_k]
\le(1-\gamma)\|z_{n,k}\|^2,\quad \gamma=\frac R{m-s}
\end{equation}
i.e., FRPG-CS satisfies~\eqref{eq:general-contraction} with $\gamma_n=R/(m-s)$.
The contraction is supplied by the reserve rather than by the trimmed mean; hence,~\eqref{eq:contraction} remains valid even when the Byzantine workers manipulate all ranking decisions, and the analysis requires neither $\widehat b\ge B$ nor an honest majority, although the numerical tests set $\widehat b=B$.
Note that the reserve is necessary, since the shared top rows alone provide no uniform contraction: when the reports of the other workers rank the first row at the top while the residual of an honest worker is supported on the second row, selecting only the first row leaves the tracking error of that worker untouched.

\subsection{Lazy Uploads (LAG-FRPG)}\label{sec:lazy}
The LAG-FRPG algorithm keeps the same recursions and state updates, but lets each honest worker upload the full residual only when the residual is large, i.e.,
\begin{equation}\label{eq:lazy}
c_{n,k}=\begin{cases}z_{n,k},&\|z_{n,k}\|^2>\tau_k,\\0,&\text{otherwise}\end{cases}
\end{equation}
with the threshold
\begin{equation}\label{eq:tau}
\tau_k=\frac{\tau_0\lambda^2G}{(k+2)\ln(k+2)},\qquad \tau_0>0.
\end{equation}

An upload of honest worker $n$ resets its state to $g_{n,k}$, while an inactive worker sends no vector and the server interprets the absence as a zero increment.
Therefore, the tracking error of each honest worker satisfies the deterministic bound
\begin{equation}\label{eq:lagerror}
\|\gh_{n,k}-g_{n,k}\|^2\le\tau_k,\qquad n=1,\ldots,N.
\end{equation}

Note that the server still broadcasts the model and each honest worker still evaluates a stochastic gradient per slot; hence, the lazy uploads reduce the uplink traffic rather than the downlink traffic or the oracle calls.

\section{Convergence Analysis}\label{sec:convergence}
\subsection{Preliminary Results}
Let $\mathcal F_k$ denote the $\sigma$-field generated by the history before the stochastic gradients of the honest workers are sampled in slot $k$, such that the points $\{u_{n,k}\}$ are $\mathcal F_k$-measurable.
The noise of the stochastic gradient of honest worker $n$ is
\begin{equation}\label{eq:noise}
\Delta_{n,k}:=\nabla f(u_{n,k};\xi_{n,k})-\nabla f_n(u_{n,k}),\quad n=1,\ldots,N.
\end{equation}

\begin{assumption}[Stochastic gradient noise]\label{as:noise}
The noise satisfies $\E[\Delta_{n,k} | \mathcal F_k]=0$ and $\E[\|\Delta_{n,k}\|^2 | \mathcal F_k] \le \sigma_n^2 <\infty$ for $n=1,\ldots,N$.
\end{assumption}

Assumption~\ref{as:noise} bounds the variance of the noise rather than the stochastic gradients, and holds for the ridge-regularized softmax regression in Section~\ref{sec:tests} with bounded input features, since the ridge term cancels in~\eqref{eq:noise}.
Analogous to~\eqref{eq:noise}, we define the deviation of the aggregated state from the sum of the penalty gradients of the honest workers as
\begin{equation}\label{eq:dev}
\Delta_{0,k}:=S_k-\sum_{n=1}^Ng_{n,k}=B_k+E_k
\end{equation}
where $B_k:=\sum_{n=N+1}^Q\gh_{n,k}$ collects the states of the Byzantine workers and $E_k:=\sum_{n=1}^N(\gh_{n,k}-g_{n,k})$ is the tracking error of the honest workers.
Based on the projection in~\eqref{eq:state}, we have for each sample path
\begin{equation}\label{eq:devbounds}
\begin{gathered}
\|S_k\|\le Q\lambda\sqrt G,\qquad
\|B_k\|\le B\lambda\sqrt G,\\
\|E_k\|\le2N\lambda\sqrt G.
\end{gathered}
\end{equation}

Different from the noise in~\eqref{eq:noise}, the deviation $\Delta_{0,k}$ can be biased and can depend on the current samples of the honest workers.
We measure the learning error by the expected objective gap $D_k:=\E[F(w_k)]-F(u^\ast)$ with $w_k:=\vecop(w_{0,k},\ldots,w_{N,k})$.
In the sequel, Assumptions~\ref{as:lipschitz}--\ref{as:noise} are satisfied, the stepsizes follow~\eqref{eq:beta} and~\eqref{eq:alpha}, all constants are independent of the horizon $K$, and all proofs are deferred to the appendices.

\begin{lemma}[One-step recursion]\label{lem:recursion}
For any $\epsilon>0$, let
\begin{align*}
\eta_{1,0,k}&:=\tfrac12(\epsilon\beta_k+\alpha_{0,k}\beta_k^2)\|u_0^\ast-v_{0,k-1}\|^2\\
&\quad-\tfrac12(\delta_0\beta_k+\alpha_{0,k}\beta_k^2)\|u_0^\ast-v_{0,k}\|^2\\
\eta_{1,n,k}&:=\tfrac12\alpha_{n,k}\beta_k^2\|u_n^\ast-v_{n,k-1}\|^2\\
&\quad-\tfrac12(\delta_n\beta_k+\alpha_{n,k}\beta_k^2)\|u_n^\ast-v_{n,k}\|^2,\quad n\ge1.
\end{align*}

Then, we have
\begin{equation}\label{eq:lem3}
\begin{split}
D_k\le{}&(1-\beta_k)D_{k-1}+\sum_{n=0}^N\E\eta_{1,n,k}\\
&+\frac{3\E(\|S_k\|+\|\Delta_{0,k}\|)^2}{2(2\alpha_{0,k}-3L_0)}
+\sum_{n=1}^N\frac{\sigma_n^2}{2(\alpha_{n,k}-L_n)}\\
&+\frac{2\E\|S_k\|^2}{\alpha_{0,k}}
+\frac{\beta_k\E\|\Delta_{0,k}\|^2}{2\epsilon}.
\end{split}
\end{equation}
\end{lemma}

Lemma~\ref{lem:recursion} extends the one-step analysis of FRPG~\cite{dong2020frpg} to an arbitrary deviation $\Delta_{0,k}$: since the noise is only cancelled against the $\mathcal F_k$-measurable $u_n^\ast-v_{n,k-1}$, neither the unbiasedness of $\Delta_{0,k}$ nor its independence of the current samples is required, and the result remains valid under adaptive attacks.

\begin{lemma}[Stability]\label{lem:bounded}
There exist finite constants $\{V_n\}_{n=0}^N$ given in~\eqref{eq:Vn} such that
\begin{equation}\label{eq:lem4}
\E\|w_{n,k}-u_n^\ast\|^2\le V_n,\qquad
\E\|v_{n,k}-u_n^\ast\|^2\le V_n
\end{equation}
for $n=0,\ldots,N$ and $k\ge0$.
\end{lemma}

\begin{lemma}[Penalty-gradient drift]\label{lem:drift}
With $g_{n,0}:=\lambda\nabla p_n(w_{0,0}-w_{n,0})$, there exist finite constants $\{\chi_n\}_{n=1}^N$ given in~\eqref{eq:cn} such that
\begin{equation}\label{eq:lem5}
\E\|g_{n,k}-g_{n,k-1}\|^2\le\chi_n/(k+2)^2.
\end{equation}
\end{lemma}

Lemma~\ref{lem:bounded} follows from~\eqref{eq:lem3} with the coarse bound $\|\Delta_{0,k}\|\le(Q+N)\lambda\sqrt G$ implied by~\eqref{eq:devbounds}; hence, the stability holds for any communication format and is not assumed by the analysis of the tracking error.
Lemma~\ref{lem:drift} shows that the penalty gradients are slowly varying signals, since the models move by ${\cal O}(1/k)$ per slot and the Lipschitz continuity of $\nabla p_n$ transfers the movement to $g_{n,k}$.

\begin{lemma}[Bounded-state tracking]\label{lem:ef}
Suppose~\eqref{eq:general-contraction} is satisfied.
Let $J_n:=\lceil8/\gamma_n\rceil$ and
\begin{equation}\label{eq:tracking-constant}
\mathcal K_n:=4(1-\gamma_n)\max\left\{
\lambda^2G(J_n+2)^2,\frac{(2-\gamma_n)\chi_n}{\gamma_n^2}\right\}.
\end{equation}

For $k\ge1$, the tracking error of honest worker $n$ satisfies
\begin{equation}\label{eq:lem6}
\zeta_{n,k}:=\E\|\gh_{n,k}-g_{n,k}\|^2
\le\mathcal K_n/(k+2)^2.
\end{equation}

In particular, $\mathcal K_n\le(1-\gamma_n)\gamma_n^{-2}
\max\{484\lambda^2G,8\chi_n\}$, and $\mathcal K_n=0$ for $\gamma_n=1$.
\end{lemma}

The proof of Lemma~\ref{lem:ef} specializes the tracking recurrence of EF21~\cite[Suppl., Lemma~2]{richtarik2021} to the drift in Lemma~\ref{lem:drift}, and combines it with the cap $\zeta_{n,k}\le4(1-\gamma_n)\lambda^2G$ implied by the bounded states in~\eqref{eq:state}, which handles the first $J_n$ slots.
The inverse-quadratic dependence of $\mathcal K_n$ on the contraction factor agrees with that of error feedback~\cite{karimireddy2019ef}, and cannot be improved uniformly over the bounded signals with the drift~\eqref{eq:lem5} and the contractive compressors (cf. Lemma~\ref{lem:tracking-sharp} in Appendix~\ref{app:tracking}).

\subsection{Main Results}
Based on Lemmas~\ref{lem:recursion}--\ref{lem:ef}, we are ready to establish the convergence of the proposed framework with the constants
\begin{align}
\Lambda_0={}&D_0+\left(\tfrac{3\delta_0}{8}+\tfrac{\alpha_{0,1}}2\right)\|u_0^\ast-v_{0,0}\|^2\nonumber\\
&+\sum_{n=1}^N\tfrac{\alpha_{n,1}}2\|u_n^\ast-v_{n,0}\|^2\label{eq:Lambda}\\
\eta(\theta)={}&\left[7Q^2+\tfrac{21}{8}(1+\theta)(Q+B)^2\right]\frac{\lambda^2G}{\delta_0}
+\sum_{n=1}^N\frac{7\sigma_n^2}{12\delta_n}.\label{eq:eta}
\end{align}

\begin{theorem}\label{thm:cs}
Suppose Assumptions~\ref{as:lipschitz}--\ref{as:noise} are satisfied and the stepsizes follow~\eqref{eq:beta} and~\eqref{eq:alpha}.
For any fixed $\theta>0$ and $K\ge1$, the proposed framework achieves
\begin{equation}\label{eq:thm1}
\begin{split}
D_K\le{}&\frac{4\Lambda_0+4K\eta(\theta)}{(K+2)^2}\\
&+(1+\theta)\frac{K^2+5K}{(K+2)^2}\frac{\lambda^2B^2G}{\delta_0}
+C_K
\end{split}
\end{equation}
where $C_K:=\frac{1+1/\theta}{2\delta_0(K+2)^2}\sum_{k=1}^K(4k+29)\E\|E_k\|^2$.
When the compressor of each honest worker satisfies~\eqref{eq:general-contraction}, we have
\begin{equation}\label{eq:CK}
C_K\le\frac{(1+1/\theta)(\sum_{n=1}^N\sqrt{\mathcal K_n})^2}
{2\delta_0(K+2)^2}
\left[4\ln\frac{K+2}{2}+21c_2\right]
\end{equation}
where $c_2:=\pi^2/6-5/4$.
\end{theorem}

Theorem~\ref{thm:cs} applies to FRPG-CS with $\gamma_n=R/(m-s)$ as well as to the individual and uniform sparsification with $\gamma_n=r/m$.
The RHS of~\eqref{eq:thm1} consists of the optimization error, the Byzantine error floor, and the compression error.
The optimization error decays as ${\cal O}(1/K)$, where $\Lambda_0$ depends on the initialization and $\eta(\theta)$ collects the impacts of the bounded states and the stochastic gradient noise; when $B=0$, $D_K$ converges to zero at the rate of ${\cal O}(1/K)$.

The Byzantine error floor is obtained by taking the limit of the horizon for a fixed $\theta$ and then the infimum over $\theta$, i.e.,
\begin{equation}\label{eq:limsup}
\limsup_{K\to\infty}D_K\le\lambda^2B^2G/\delta_0.
\end{equation}

Note that the finite-time bound~\eqref{eq:thm1} holds for each fixed $\theta$ and is not uniform as $\theta\downarrow0$.
The floor in~\eqref{eq:limsup} is independent of the compressor, i.e., the compression does not degrade the Byzantine resilience of the penalty-based method.
Moreover, the floor grows with $\lambda^2$ and $B^2$ without requiring an honest majority; hence, a smaller penalty weight improves the Byzantine resilience at the expense of a larger model bias (cf. Remark~\ref{rem:bias}).

The compression error $C_K$ decays as ${\cal O}(\ln K/K^2)$, which is faster than the optimization error, and depends on the compressors only through the constants $\{\mathcal K_n\}$ that scale as $(1-\gamma_n)/\gamma_n^2$ and vanish at full communication.
Therefore, a smaller contraction factor reduces the uplink traffic per slot at the expense of a larger constant of a higher-order term.

Since Theorem~\ref{thm:cs} bounds the expected objective gap of~\eqref{eq:penalized}, the strong convexity implies $\E\|w_K-u^\ast\|^2\le2D_K/\delta_{\min}$.
Together with Lemma~\ref{lem:consensus}, the distance to the consensus optimum is separated into the algorithmic error and the model bias as
\begin{equation}\label{eq:consensus-bridge}
\sqrt{\E\|w_{0,K}-x^\star\|^2}
\le\sqrt{2D_K/\delta_0}+\|u_0^\ast-x^\star\|
\end{equation}
where the second term on the RHS is not removed by the convergence of the algorithm.

\begin{theorem}\label{thm:lazy}
Under the conditions of Theorem~\ref{thm:cs}, the LAG-FRPG algorithm with the threshold~\eqref{eq:tau} achieves~\eqref{eq:thm1} with
\begin{equation}\label{eq:thm2}
C_K\le\frac{11(1+1/\theta)N^2\tau_0\lambda^2G}
{\delta_0(K+2)\ln(K+2)}.
\end{equation}

Consequently, the Byzantine error floor~\eqref{eq:limsup} also holds for LAG-FRPG.
\end{theorem}

Theorem~\ref{thm:lazy} follows from the deterministic bound~\eqref{eq:lagerror}, which gives $\|E_k\|\le N\sqrt{\tau_k}$.
The lazy uploads add an ${\cal O}(1/(K\ln K))$ term that is dominated by the optimization error, while the Byzantine error floor remains unchanged.
The number of uploads is characterized in the following corollary.

\begin{corollary}[Number of lazy uploads]\label{cor:uploads}
Let $U_{n,K}:=\sum_{k=1}^KI_{n,k}$ denote the number of uploads of honest worker $n$ during slots $1,\ldots,K$, where $I_{n,k}\in\{0,1\}$ indicates an upload in slot $k$.
Under the conditions of Theorem~\ref{thm:cs}, for any positive and nonincreasing thresholds $\{\tau_k\}$ with $\omega_k:=\sqrt{\tau_k}$, we have
\begin{equation}\label{eq:normalized-count}
\E U_{n,K}\le1+\sqrt{\chi_n}\sum_{k=2}^K\frac1{(k+2)\omega_k}
+\sum_{k=2}^K\left(\frac{\omega_{k-1}}{\omega_k}-1\right).
\end{equation}

In particular, the threshold~\eqref{eq:tau} yields
\begin{equation}\label{eq:prop1}
\begin{split}
\E U_{n,K}\le{}&1+2\sqrt{\chi_n/\bar\tau}\sqrt{\ln M}(\sqrt M-\sqrt3)\\
&+\ln((K+1)/2)
\end{split}
\end{equation}
where $\bar\tau:=\tau_0\lambda^2G$ and $M:=K+2$.
\end{corollary}

Corollary~\ref{cor:uploads} shows that the expected fraction of uploading slots of LAG-FRPG decays as ${\cal O}(\sqrt{\ln K/K})$.
The proof relies on the normalized tracking error $\|g_{n,k}-\gh_{n,k}\|/\omega_k\in[0,1]$, which is summed along each sample path before taking the expectation; compared with the bound ${\cal O}((\ln K)^{3/2}/\sqrt K)$ that only uses the final threshold and the total path length of the penalty gradients, such an argument removes a logarithmic factor.
Note that the threshold controls the tradeoff between the tracking accuracy and the upload frequency, e.g., the alternative threshold $\tau_k=\bar\tau\ln^2(k+2)/(k+2)^2$ reduces the additional objective error to ${\cal O}(\ln^3K/K^2)$, while the expected fraction of uploading slots only vanishes as ${\cal O}(1/\ln K)$ (cf. Appendix~\ref{app:uploads}).

\section{Numerical Results}\label{sec:tests}

\subsection{Experimental Setup}\label{sec:setup}
\subsubsection{\bfseries Datasets and Models}
We evaluate the proposed algorithms on the MNIST~\cite{mnist} and Fashion-MNIST~\cite{xiao2017fashion} datasets with the ten-class linear softmax regression, which has $785\times10=7{,}850$ parameters with a constant input feature representing the bias.
All parameters of the honest workers are regularized by a ridge term with weight $\rho=0.01$, and the server regularizer is a ridge term with weight $\rho_0=0.1$; hence, Assumptions~\ref{as:lipschitz} and~\ref{as:strong} are satisfied.

\subsubsection{\bfseries Federated Settings}
We set $N=10$ honest workers and add $B=2$ Byzantine workers unless otherwise specified, without changing the data of the honest workers.
A fixed stratified split of the training set yields 50,000 training and 10,000 validation examples, and the 10,000 test examples are only used for testing.
The training examples are partitioned among the honest workers by a label-Dirichlet distribution with concentration $a=0.3$, where a partition is redrawn when a worker receives fewer than 100 examples.
For each seed, all methods share the same partition and mini-batch indices with 32 examples sampled with replacement per worker and slot.
All methods start from the zero model and run for 2,000 oracle slots in double precision, where each local step of LFRPG is counted as one slot, i.e., each run consumes 640,000 sample-gradient evaluations.

We report three sets of runs.
The main comparison uses 20 seeds (100--119) in the attack-free and Min-Max scenarios.
An exploratory study uses five seeds (0--4) over eight scenarios, including the IPM, ALIE, and Gaussian-noise attacks, $a\in\{0.1,10\}$, and $B=4$, together with the ablations in Section~\ref{sec:ablation}; since the main comparison was designed after inspecting the exploratory study, the two seed sets are reported separately.
A controlled study fixes $(\lambda,\mu)=(0.5,0.1)$, eight transmitted rows, and $\tau_0=1$ with all 60,000 training examples for the diagnostics in Sections~\ref{sec:ablation} and~\ref{sec:bias}.

\subsubsection{\bfseries Benchmarks}
We compare FRPG-CS and LAG-FRPG with ten benchmarks, all of which are implemented by ourselves:
FRPG and LFRPG with frame length $T=20$~\cite{dong2020frpg},
projected FRPG (i.e., the proposed framework with full uploads),
$\ell_2$-norm RSA~\cite{li2019rsa},
trimmed mean and coordinate-wise median~\cite{yin2018},
Krum~\cite{blanchard2017},
geometric median~\cite{pillutla2022rfa},
Byz-EF21-SGDM~\cite{liu2026momentum}, and
FedAdam-BACK~\cite{xu2026fedadamback}.
The implementation details of the benchmarks are provided in Appendix~\ref{app:implementation}.

\subsubsection{\bfseries Hyperparameter Tuning Protocol}
For a fair comparison, each method receives six candidate configurations, which are scored by the mean validation cross-entropy after 1,000 slots over the attack-free, IPM, and ALIE scenarios with the calibration seed 9001.
The selected configurations are frozen before the evaluation; hence, the Min-Max and Gaussian-noise attacks, the alternative concentrations, and the case of $B=4$ are not used for tuning.
The tuning selects $(\lambda,\mu,s+R)=(0.5,0.1,16)$ with $\ell=50$, $s=13$, and $R=3$ for FRPG-CS, $(\lambda,\mu,\tau_0)=(0.5,0.1,100)$ for LAG-FRPG, $(\lambda,\mu)=(0.5,0.05)$ for projected FRPG, and $\lambda=0.1$ for FRPG and LFRPG.
Since the selected configurations optimize different finite-penalty objectives and several selected values lie on the boundary of the candidate grids, we complement the main comparison with a matched-objective comparison in Section~\ref{sec:matched} and the ablations in Section~\ref{sec:ablation}.

\subsubsection{\bfseries Malicious Attacks}
The Byzantine workers collude and observe the vectors $\{y_n\}_{n=1}^N$ that the considered protocol aggregates from the honest workers, i.e., the penalty gradients, the (reconstructed) momentum, or the loss gradients, with the mean $\bar y$.
They submit $-10\bar y$ in the IPM attack~\cite{xie2019ipm}, $\bar y-1.5\sigma_y$ with $\sigma_y$ being the coordinate-wise standard deviation of the honest vectors in the ALIE-type attack~\cite{baruch2019alie}, and a Gaussian vector whose norm is ten times the root-mean-square norm of the honest vectors in the Gaussian-noise attack.
In the Min-Max attack~\cite{shejwalkar2021minmax}, they submit $y^{\rm B}:=\bar y-\varsigma\bar y/\|\bar y\|$ with the largest $\varsigma\ge0$ that satisfies
\begin{equation}\label{eq:benchmark-minmax}
\max_n\|y^{\rm B}-y_n\|^2\le\max_{i,j}\|y_i-y_j\|^2
\end{equation}
where a fixed unit direction replaces $\bar y/\|\bar y\|$ when $\bar y=0$.
For the protocols with states, the targets are converted into increments in the admissible sparse format, and the Byzantine workers in FRPG-CS additionally forge the energy reports on the rows with low honest energy before the reserve is revealed.
Since the feasible payloads differ across the protocols, the attacks share the same information access but not necessarily the same strength for all methods.
Note that the IPM and Gaussian-noise payloads exceed the norm bound required by FRPG and LFRPG; hence, the degradation of FRPG and LFRPG under such attacks does not contradict the guarantees in~\cite{dong2020frpg}.

\subsubsection{\bfseries Performance Metrics}
We report the test accuracy at 2,000 slots and at the budget endpoints, i.e., the last models before the uplink traffic averaged over the honest workers exceeds 1, 4, or 8~MiB, where the traffic counts 64-bit scalars and 32-bit indices including the energy reports, the masks, and the model downloads per recipient.
The methods are compared by paired $t$-tests across the seeds with the Holm correction, and the reported intervals are marginal 95\% confidence intervals conditional on the fixed test images.

\subsection{Performance Comparison}\label{sec:main-comparison}
\begin{table*}[!t]
\centering
\caption{Test accuracy (\%, mean $\pm$ standard deviation over 20 seeds) with $a=0.3$ and $B=2$. Clean and Min-Max: 2,000 slots; 8~MiB: Min-Max scenario under the budget of 8~MiB on the mean uplink traffic of the honest workers. M: MNIST; F: Fashion-MNIST. Bold: highest mean per column.}
\label{tab:confirmation-accuracy}
\setlength{\tabcolsep}{4pt}
\begin{tabular}{lcccccc}
\toprule
Method & M: Clean & M: Min-Max & M: 8 MiB & F: Clean & F: Min-Max & F: 8 MiB \\
\midrule
FRPG-CS (ours) & $88.60\pm0.57$ & $\bestresult{88.46\pm0.66}$ & $\bestresult{88.43\pm0.82}$ & $78.48\pm1.11$ & $\bestresult{78.38\pm1.09}$ & $\bestresult{78.07\pm1.06}$ \\
LAG-FRPG (ours) & $88.72\pm0.61$ & $87.90\pm0.76$ & $88.26\pm0.68$ & $78.57\pm1.09$ & $77.02\pm1.31$ & $77.75\pm1.12$ \\
FRPG & $86.49\pm1.98$ & $85.47\pm2.41$ & $82.11\pm3.48$ & $75.54\pm1.99$ & $73.87\pm2.40$ & $66.55\pm3.39$ \\
Projected FRPG & $88.68\pm0.65$ & $85.97\pm1.55$ & $85.51\pm1.66$ & $78.49\pm1.12$ & $73.99\pm1.80$ & $72.01\pm1.90$ \\
LFRPG & $86.30\pm2.01$ & $85.28\pm2.45$ & $85.28\pm2.45$ & $75.19\pm2.06$ & $73.60\pm2.42$ & $73.60\pm2.42$ \\
$\ell_2$-RSA & $88.49\pm0.70$ & $85.81\pm1.66$ & $79.80\pm3.15$ & $78.18\pm1.18$ & $73.68\pm1.87$ & $67.17\pm2.92$ \\
Trimmed mean & $89.40\pm0.17$ & $81.10\pm4.44$ & $72.90\pm5.73$ & $76.84\pm2.35$ & $70.30\pm3.69$ & $61.79\pm5.11$ \\
Coord. median & $81.33\pm3.34$ & $65.73\pm7.67$ & $57.60\pm7.56$ & $72.36\pm3.81$ & $63.59\pm6.91$ & $51.85\pm6.13$ \\
Krum & $80.37\pm4.51$ & $65.93\pm7.82$ & $57.14\pm5.16$ & $68.18\pm5.69$ & $59.93\pm6.95$ & $58.53\pm7.41$ \\
Geometric median & $88.92\pm0.47$ & $84.89\pm2.46$ & $80.77\pm2.82$ & $79.51\pm1.07$ & $74.68\pm2.39$ & $62.10\pm4.78$ \\
Byz-EF21-SGDM & $\bestresult{89.65\pm0.19}$ & $81.41\pm4.00$ & $81.68\pm3.94$ & $\bestresult{80.75\pm0.48}$ & $67.76\pm4.11$ & $67.70\pm4.43$ \\
FedAdam-BACK & $84.77\pm1.49$ & $73.43\pm4.17$ & $73.38\pm3.19$ & $74.37\pm1.52$ & $67.63\pm3.57$ & $68.03\pm3.24$ \\
\bottomrule
\end{tabular}
\end{table*}

\begin{figure*}[!t]
\centering
\includegraphics[width=5.0in]{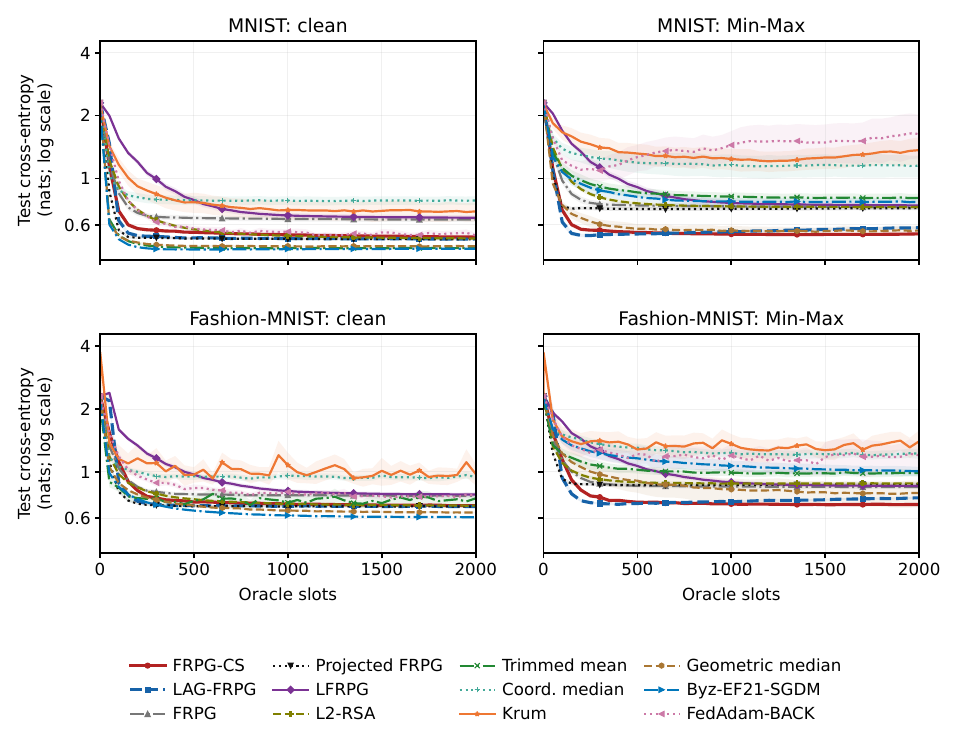}
\caption{Test cross-entropy (log scale) versus oracle slots for all twelve methods in the attack-free and Min-Max ($B=2$) scenarios. Lines and shaded regions show the means and the pointwise 95\% confidence intervals over 20 seeds.}
\label{fig:convergence-followup}
\end{figure*}

Table~\ref{tab:confirmation-accuracy} summarizes the test accuracy in the main comparison.
Under the Min-Max attack, FRPG-CS and LAG-FRPG achieve the highest and the second-highest accuracy on both datasets at 2,000 slots as well as under the 8-MiB uplink budget.
Compared with the best benchmark at 2,000 slots, i.e., projected FRPG on MNIST and geometric median on Fashion-MNIST, FRPG-CS improves the accuracy by 2.49 (95\% confidence interval $[2.04,2.93]$) and 3.70 ($[2.63,4.77]$) percentage points, respectively, and LAG-FRPG by 1.93 ($[1.51,2.34]$) and 2.34 ($[1.19,3.49]$) percentage points, respectively, where the Holm-adjusted two-sided $p$-values are below 0.001 except for the last one (0.006).
More generally, among the 160 paired contrasts between the two proposed algorithms and the ten benchmarks over two datasets, two scenarios, and two endpoints, each proposed algorithm outperforms each benchmark in all 40 Min-Max contrasts after the correction; note that such contrasts are not independent replications.
In the attack-free scenario, Byz-EF21-SGDM achieves the highest accuracy on both datasets, i.e., 89.65\% and 80.75\%, and outperforms both proposed algorithms.

Fig.~\ref{fig:convergence-followup} shows the test cross-entropy versus the number of oracle slots, which is a common predictive metric rather than the training objective of each method or the objective gap in Theorem~\ref{thm:cs}.
All methods start from the zero model with the test cross-entropy of $\ln10$.
Under the Min-Max attack, LAG-FRPG has a lower mean loss than FRPG-CS at slot 500 on both datasets but a higher mean loss at slot 2,000, while Byz-EF21-SGDM attains the lowest final mean loss in the attack-free scenario.

\subsection{Comparison of Communication Formats}\label{sec:matched}
To remove the confound of different finite-penalty objectives, we fix $\lambda=0.5$, $\mu=0.1$, the stepsizes, the partitions, and the mini-batch streams, and compare four communication formats of the proposed framework: the full upload, the shared mask of FRPG-CS with $s=13$ and $R=3$, the individual top-16 rows, and 16 uniformly sampled common rows.
The 320 runs cover both datasets, the attack-free and Min-Max scenarios, and the seeds 100--119 without retuning; since they reuse the partitions and the test sets of the main comparison, they do not serve as an independent confirmation.

\begin{table*}[!t]
\centering\setlength{\tabcolsep}{3pt}
\caption{Test accuracy (\%, mean $\pm$ standard deviation over 20 seeds) of four communication formats with the matched objective ($\lambda=0.5$, $\mu=0.1$, and 16 rows for the sparse formats). The columns are defined as in Table~\ref{tab:confirmation-accuracy}. Bold: highest mean per column.}
\label{tab:matched-compression}
\begin{tabular}{lrrrrrr}
\toprule
Method & M: Clean & M: Min-Max & M: 8 MiB & F: Clean & F: Min-Max & F: 8 MiB\\
\midrule
Projected full & $\bestresult{88.72\pm0.63}$ & $85.98\pm1.56$ & $85.50\pm1.66$ & $78.55\pm1.09$ & $73.99\pm1.80$ & $72.02\pm1.90$ \\
Shared reserve & $88.60\pm0.57$ & $\bestresult{88.46\pm0.66}$ & $\bestresult{88.43\pm0.82}$ & $78.48\pm1.11$ & $\bestresult{78.38\pm1.09}$ & $78.07\pm1.06$ \\
Individual top & $88.68\pm0.61$ & $87.28\pm1.29$ & $88.11\pm0.84$ & $\bestresult{78.62\pm0.92}$ & $78.28\pm1.04$ & $\bestresult{78.18\pm1.07}$ \\
Uniform common & $88.32\pm0.68$ & $88.29\pm0.70$ & $88.07\pm0.76$ & $78.14\pm1.19$ & $78.11\pm1.22$ & $77.74\pm1.25$ \\
\bottomrule
\end{tabular}
\end{table*}

Table~\ref{tab:matched-compression} summarizes the test accuracy, where the 24 contrasts between the shared mask and the other three formats form the family for the Holm correction.
Under the Min-Max attack, the shared mask outperforms the full upload on both datasets at both endpoints.
Compared with the individual top rows on MNIST, the shared mask improves the accuracy by 1.18 percentage points at 2,000 slots (95\% confidence interval $[0.85,1.51]$, adjusted $p=7.43\times10^{-6}$) and by 0.32 percentage points under the 8-MiB budget ($p=0.0043$), while the corresponding contrasts on Fashion-MNIST remain unresolved.
Compared with the uniformly sampled common rows, the two contrasts at 2,000 slots under the Min-Max attack remain unresolved, while the contrast under the 8-MiB budget on MNIST favors the shared mask by 0.36 percentage points ($p=0.0027$).
In the attack-free scenario, the contrasts with the full upload and the individual top rows are all unresolved, with slightly lower means for the shared mask.
Hence, the shared mask has conditional rather than uniform advantages over the simpler formats, all of which are covered by Theorem~\ref{thm:cs}.

At 2,000 slots, the mean uplink traffic per honest worker is 119.78~MiB for the full upload, 14.60~MiB for the shared mask, 12.33~MiB for the individual top rows, and 12.21~MiB for the uniformly sampled common rows, including the energy reports of the shared mask and the indices of the individual top rows.
In terms of the common objective~\eqref{eq:penalized}, the individual top rows attain a lower mean value than the shared mask on Fashion-MNIST under the Min-Max attack (6.204 versus 6.333) despite the slightly lower mean accuracy (cf. Table~\ref{tab:matched-objective} in Appendix~\ref{app:implementation}).

\subsection{Communication Overhead and Ablation Studies}\label{sec:ablation}
\subsubsection{\bfseries Communication Overhead}
With the selected configuration, FRPG-CS uploads $157+16\times50=957$ scalars per honest worker and slot, i.e., 12.19\% of the dense vector with 7,850 entries, such that its uplink traffic over 2,000 slots is 14.60~MiB compared with 119.78~MiB for the full upload.
Since the server broadcasts the model per slot, the total traffic of FRPG-CS including the downlink is 134.51~MiB per honest worker.
In contrast, LFRPG requires 11.98~MiB in total at the expense of a lower accuracy under the Min-Max attack, and Byz-EF21-SGDM and FedAdam-BACK also have a smaller total traffic than the proposed algorithms.
Moreover, under the strict 1-MiB uplink budget on Fashion-MNIST, LAG-FRPG trails LFRPG and several compressed benchmarks; hence, the advantage of the proposed algorithms depends on the available communication resources.

\subsubsection{\bfseries Mechanism Ablation}
\begin{table}[!t]
\centering
\caption{Ablation study: test accuracy (\%, mean $\pm$ standard deviation over five seeds) at 2,000 slots with $B=2$ and $a=0.3$. EF: error feedback. Bold: highest mean per column.}
\label{tab:benchmark-ablations}
\setlength{\tabcolsep}{3pt}
\resizebox{\columnwidth}{!}{%
\begin{tabular}{lcccc}
\toprule
Method & M: IPM & M: Min-Max & F: IPM & F: Min-Max \\
\midrule
FRPG-CS & $\bestresult{86.95\pm1.70}$ & $86.82\pm1.74$ & $\bestresult{79.25\pm1.05}$ & $\bestresult{79.35\pm1.03}$ \\
CS: no EF & $60.02\pm5.16$ & $35.95\pm3.19$ & $36.88\pm5.09$ & $12.54\pm2.51$ \\
CS: no reserve & $85.62\pm2.57$ & $83.14\pm3.88$ & $78.95\pm0.99$ & $78.96\pm0.93$ \\
CS: no projection & $2.97\pm3.07$ & $\bestresult{86.85\pm1.72}$ & $6.74\pm3.32$ & $79.30\pm1.02$ \\
LAG-FRPG & $85.84\pm1.84$ & $86.10\pm1.68$ & $78.00\pm0.99$ & $78.29\pm0.82$ \\
LAG: no projection & $3.48\pm4.35$ & $86.09\pm1.67$ & $4.31\pm4.88$ & $78.30\pm0.81$ \\
\bottomrule
\end{tabular}}
\end{table}
Table~\ref{tab:benchmark-ablations} shows the ablations of the exploratory study, where each ablation inherits the selected hyperparameters and the communication format of its parent.
Without the error feedback, i.e., when the honest workers upload the masked penalty gradients and the server overwrites the states, the accuracy of FRPG-CS drops to 60.02\% and 36.88\% under the IPM attack and to 35.95\% and 12.54\% under the Min-Max attack.
Without the projection in~\eqref{eq:state}, both proposed algorithms fail under the IPM attack with its large payloads, while the projection has little effect under the Min-Max attack.
Without the reserve rows, i.e., when all 16 rows are selected by the ranking, the mean accuracy of FRPG-CS decreases in all four cases, e.g., by 3.68 percentage points on MNIST under the Min-Max attack.
Hence, the ablations support the roles of the residual feedback and the projection in the tested scenarios, although they do not imply that each component improves the accuracy under every attack.

\subsubsection{\bfseries Objective Gap and Upload Frequency}
In the controlled study, where the minimizer of~\eqref{eq:penalized} is computed with a certified objective error below $3.43\times10^{-9}$, FRPG-CS under two sign-flipping Byzantine workers attains a higher accuracy of the server model but a larger final objective gap than the full upload on both datasets; hence, a higher accuracy does not imply a smaller objective gap.
Moreover, the individual top rows with eight rows attain an accuracy comparable to that of the shared mask with a smaller uplink traffic (cf. Appendix~\ref{app:implementation}); therefore, we do not claim that the shared mask outperforms the individual sparsification in general.
Finally, ten runs of LAG-FRPG over 20,000 slots upload in 8.08\% and 11.36\% of the slots in the final window on the two datasets, compared with the cumulative fractions of 21.82\% and 26.97\%, respectively, which agrees with the decreasing upload activity in Corollary~\ref{cor:uploads}.

\subsection{Finite-Penalty Model Bias}\label{sec:bias}
\begin{table}[!t]
\centering
\caption{Model bias $\|u_0^\ast-x^\star\|$ with $\lambda=0.5$ and $\mu=0.1$: enclosure of the exact values over five seeds and the largest upper bound from~\eqref{eq:consensus-residual-distance}. Bold: smallest values per dataset.}
\label{tab:consensus-diagnostics}
\setlength{\tabcolsep}{4pt}
\begin{tabular}{lcc}
\toprule
Data, $a$ & Distance enclosure & Max. bound \\
\midrule
M, $0.3$ & $[0.339,0.621]$ & $12.574$ \\
M, $10$ & $\bestresult{[0.023,0.035]}$ & $\bestresult{0.377}$ \\
F, $0.3$ & $[0.531,0.945]$ & $37.980$ \\
F, $10$ & $\bestresult{[0.027,0.050]}$ & $\bestresult{1.198}$ \\
\bottomrule
\end{tabular}
\end{table}
We evaluate the certificates in Lemma~\ref{lem:consensus} on the 20 reference problems of the controlled study, i.e., two datasets, $a\in\{0.3,10\}$, and five seeds, with $\lambda=0.5$ and $\mu=0.1$.
The consensus optimum and the penalized optimum are computed from the full data of the honest workers, where the numerical errors are bounded by the strong-convexity residuals with the largest half-width of the distance intervals being $8.31\times10^{-4}$.
Table~\ref{tab:consensus-diagnostics} reports the interval that encloses the exact distance $\|u_0^\ast-x^\star\|$ of all five seeds, which is deterministic rather than a confidence interval, together with the largest upper bound from the gradient certificate~\eqref{eq:consensus-residual-distance}.

With $a=0.3$, the model bias is appreciable, i.e., between 0.339 and 0.621 on MNIST and between 0.531 and 0.945 on Fashion-MNIST, while it is one order of magnitude smaller with $a=10$.
The threshold $\max_n\|\nabla f_n(x^\star)\|$ ranges from 1.047 to 1.683 on MNIST and from 2.088 to 4.536 on Fashion-MNIST with $a=0.3$, which exceeds $\lambda=0.5$; hence, $A_\lambda>0$ in~\eqref{eq:unconditional-huber-envelope}, and the positive bias is confirmed by the enclosures.
Raising the penalty weight would raise the Byzantine error floor in~\eqref{eq:limsup}, which agrees with the tradeoff in Remark~\ref{rem:bias}.
The source code, the configurations, and the raw results of all 3,012 runs are provided in the supplementary archive (cf. Appendix~\ref{app:implementation}).

\section{Concluding Remarks}\label{sec:conclusion}
We have proposed a projected penalty-gradient tracking framework for communication-efficient and Byzantine-resilient FL over heterogeneous datasets, and developed the FRPG-CS algorithm with a reserve-protected shared mask and the LAG-FRPG algorithm with event-triggered uploads.
For smooth and strongly convex losses, we have established a finite-time bound on the expected objective gap, where the Byzantine error floor $\lambda^2B^2G/\delta_0$ is independent of the compressor, the compression error decays as ${\cal O}(\ln K/K^2)$ with a tight dependence on the contraction factor, and the expected fraction of lazy uploads decays as ${\cal O}(\sqrt{\ln K/K})$; we have also derived unconditional certificates that separate the finite-penalty model bias from the optimization error.
Numerical results on MNIST and Fashion-MNIST have shown that FRPG-CS and LAG-FRPG outperform ten benchmarks under the Min-Max attack, although Byz-EF21-SGDM attains a higher attack-free accuracy and LFRPG requires a smaller total traffic.
Since our analysis requires strongly convex losses, a trusted server, and reliable links, and our evaluation is limited to linear models on two image datasets, future work will extend the framework to non-convex losses and downlink compression, and validate it on larger models and stronger protocol-specific attacks.

\ifdefined\TSPsplit
  \printreferences
  \clearpage
\fi
\appendices

\section{Proof of Lemma~\ref{lem:consensus}}\label{app:consensus}
Since the penalty gradients enter the gradients of $F$ with respect to $w_0$ and $w_n$ with opposite signs, we have
$\sum_{n=0}^N\nabla_nF(w)=\nabla f_0(w_0)+\sum_{n=1}^N\nabla f_n(w_n)$.
Therefore,
\[
\nabla H(w_0)=\sum_{n=0}^N\nabla_nF(w)+
 \sum_{n=1}^N[\nabla f_n(w_0)-\nabla f_n(w_n)]
\]
whose norm is bounded by $b(w)$ under Assumption~\ref{as:lipschitz}.
The strong monotonicity of $\nabla H$ gives~\eqref{eq:consensus-residual-distance}, and minimizing both sides of
$H(x)\ge H(w_0)+\langle\nabla H(w_0),x-w_0\rangle
+\delta_\Sigma\|x-w_0\|^2/2$ over $x$ gives~\eqref{eq:consensus-residual-objective}.
For an approximate minimizer $\bar u$ of $F$, the strong convexity of $F$ yields $\|\bar u-u^\ast\|\le\|\nabla F(\bar u)\|/\delta_{\min}$, and the triangle inequality gives the certificate stated after Lemma~\ref{lem:consensus}.
Moreover, the strong convexity of $F$ in $w_0$ yields
$\E\|w_{0,K}-u_0^\ast\|^2\le2D_K/\delta_0$, and Minkowski's
inequality in $L^2$ proves~\eqref{eq:consensus-bridge}.

To prove~\eqref{eq:unconditional-huber-envelope}, let $\check F$ denote the function obtained by replacing each Huber penalty in $F$ with the Euclidean norm, let $\check u$ denote the minimizer of $\check F$, and let $\bar x^\star:=\vecop(x^\star,\ldots,x^\star)$.
We use the block norm $\|w\|_\delta^2:=\sum_{n=0}^N\delta_n\|w_n\|^2$.
At $\bar x^\star$, we choose the subgradients $\Pi_\lambda(\nabla f_n(x^\star))$ of the norm penalties, such that a subgradient of $\check F$ at $\bar x^\star$ has the blocks
$(-\sum_{n=1}^Ne_n,e_1,\ldots,e_N)$ and the dual norm $A_\lambda$, where the first block follows from $\nabla H(x^\star)=0$.
The strong monotonicity of $\partial\check F$ and the weighted Cauchy--Schwarz inequality imply
$\|\check u-\bar x^\star\|_\delta\le A_\lambda$.
Since the Huber penalty satisfies
$0\le\check F(w)-F(w)\le N\lambda\mu/2$, the optimality of $u^\ast$ for $F$ gives
$\check F(u^\ast)-\check F(\check u)\le N\lambda\mu/2$, and the strong convexity of $\check F$ yields
$\|u^\ast-\check u\|_\delta\le\sqrt{N\lambda\mu}$.
The triangle inequality and
$\sqrt{\delta_0}\|u_0^\ast-x^\star\|\le\|u^\ast-\bar x^\star\|_\delta$
prove~\eqref{eq:unconditional-huber-envelope}.
When only a numerical consensus solution $\bar x$ is available, $A_\lambda$ is replaced by the dual norm of the subgradient with the blocks
$\nabla f_0(\bar x)+\sum_{n=1}^N\Pi_\lambda(\nabla f_n(\bar x))$
and $\nabla f_n(\bar x)-\Pi_\lambda(\nabla f_n(\bar x))$, and $\|\nabla H(\bar x)\|/\delta_\Sigma$ is added to the RHS of~\eqref{eq:unconditional-huber-envelope}, which is fully computable.

\subsection{Profiled Loss}
Eliminating the local models from~\eqref{eq:penalized} gives the profiled loss
\begin{align*}
\Phi(x)&:=f_0(x)+\sum_{n=1}^N\phi_n(x)\\
\phi_n(x)&:=\inf_y\{f_n(y)+\lambda p(x-y)\}.
\end{align*}

The inner minimizer $y_n(x)$ exists and is unique by the strong convexity and the coercivity in $y$, the partial minimization preserves the convexity, and the envelope formula gives
$\nabla\phi_n(x)=\lambda\nabla p(x-y_n(x))$ with the norm of at most $\lambda$.
Hence, $\Phi$ is $\delta_0$-strongly convex with the unique minimizer $u_0^\ast$, i.e., the finite-penalty target is a profiled loss with bounded gradient contributions of the workers.

\subsection{Conditional Bound in Remark~\ref{rem:bias}}
If $\lambda>\max_n\|\nabla f_n(u_n^\ast)\|$, the stationarity implies
$u_0^\ast-u_n^\ast=\mu\nabla f_n(u_n^\ast)/\lambda$ and hence
$\|u_0^\ast-u_n^\ast\|<\mu$.
Summing the stationarity conditions gives $\|\nabla H(u_0^\ast)\|\le\mu\sum_{n=1}^NL_n$, and the strong monotonicity of $\nabla H$ yields the bound in Remark~\ref{rem:bias}.
The condition can fail: for $f_0(x)=x^2/2$, $f_1(x)=(x-4)^2/2$, and $\lambda=1$,
the penalized optimum is $(1,3)$ for any $0<\mu<2$, whereas $x^\star=2$.

\section{Proof of Lemma~\ref{lem:recursion}}\label{app:proofs}
In Appendices~\ref{app:proofs}--\ref{app:uploads}, the initialization is deterministic and finite, all messages are finite real vectors, and the constants are only used in the analysis and need not be known to the algorithm.
In this appendix, we abbreviate $\alpha_{n,k}$ and $\beta_k$ by $\alpha_n$ and $\beta$, and let $h_0:=\nabla f_0(u_{0,k})$ and $h_n:=\nabla f(u_{n,k};\xi_{n,k})-g_{n,k}$ for $n\ge1$.
Based on~\eqref{eq:server-w} and the optimality condition of the proximal operator in~\eqref{eq:worker-w}, we have $\alpha_n(u_{n,k}-w_{n,k})=h_n$ for $n=0,\ldots,N$.

For any comparison point $x=\vecop(x_0,\ldots,x_N)$, the smoothness and the strong convexity in Assumptions~\ref{as:lipschitz} and~\ref{as:strong} give
\begin{align*}
&f_0(w_{0,k})-f_0(x_0)\\
&\quad\le\langle S_k,x_0-w_{0,k}\rangle
 +\frac{\|S_k\|\|h_0\|}{\alpha_0}\\
&\qquad-\frac{2\alpha_0-L_0}{2\alpha_0^2}\|h_0\|^2
 -\frac{\delta_0}{2}\|x_0-u_{0,k}\|^2\\
&\qquad-\langle h_0+S_k,x_0-u_{0,k}\rangle\\
&f_n(w_{n,k})-f_n(x_n)\\
&\quad\le\langle\Delta_{n,k}-g_{n,k},x_n-w_{n,k}\rangle
 -\frac{2\alpha_n-L_n}{2\alpha_n^2}\|h_n\|^2\\
&\qquad-\frac{\delta_n}{2}\|x_n-u_{n,k}\|^2
 -\langle h_n,x_n-u_{n,k}\rangle
\end{align*}
where the second inequality is obtained by starting from
$f_n(x_n)\ge f_n(u_{n,k})+\langle\nabla f_n(u_{n,k}),x_n-u_{n,k}\rangle
+\delta_n\|x_n-u_{n,k}\|^2/2$,
applying the smoothness upper bound to $f_n(w_{n,k})$, and using
$\nabla f_n(u_{n,k})=h_n+g_{n,k}-\Delta_{n,k}$; the first inequality is obtained similarly by adding and subtracting $S_k$ and using
$\langle S_k,h_0\rangle\ge-\|S_k\|\|h_0\|$.
Moreover, the convexity of the penalty function gives
\begin{align*}
&\lambda[p_n(w_{0,k}-w_{n,k})-p_n(x_0-x_n)]\\
&\quad\le\langle g_{n,k},w_{0,k}-x_0+x_n-w_{n,k}\rangle.
\end{align*}

Let $\tilde h_0:=h_0+S_k$, $\tilde h_n:=h_n$ for $n\ge1$, and
\[
\Omega_k:=\frac{\|S_k\|\|h_0\|}{\alpha_0}
 -\sum_{n=0}^N\frac{2\alpha_n-L_n}{2\alpha_n^2}\|h_n\|^2.
\]

Summing the above inequalities over the server and the honest workers, the penalty gradients of the honest workers cancel, and we obtain
\begin{equation}\label{eq:supp-summed}
\begin{split}
F(w_k)-F(x)\le{}&\sum_{n=0}^N\langle\Delta_{n,k},x_n-w_{n,k}\rangle+\Omega_k\\
&-\sum_{n=0}^N\Bigl[\frac{\delta_n}{2}\|x_n-u_{n,k}\|^2\\
&\qquad+\langle\tilde h_n,x_n-u_{n,k}\rangle\Bigr]
\end{split}
\end{equation}
where $\Delta_{0,k}$ is the deviation in~\eqref{eq:dev}.

The updates~\eqref{eq:server-v} and~\eqref{eq:worker-v} imply that $v_{n,k}$ minimizes the quadratic function
$\langle\tilde h_n,x-u_{n,k}\rangle+
\delta_n\|x-u_{n,k}\|^2/2+
\alpha_n\beta\|x-v_{n,k-1}\|^2/2$.
After discarding $\delta_n\|v_{n,k}-u_{n,k}\|^2/2\ge0$, the optimality of $v_{n,k}$ gives
\begin{align*}
&\langle\tilde h_n,x-u_{n,k}\rangle+\frac{\delta_n}{2}\|x-u_{n,k}\|^2\\
&\quad\ge\langle\tilde h_n,v_{n,k}-u_{n,k}\rangle
 +\frac{\alpha_n\beta+\delta_n}{2}\|x-v_{n,k}\|^2\\
&\qquad+\frac{\alpha_n\beta}{2}\|v_{n,k}-v_{n,k-1}\|^2
 -\frac{\alpha_n\beta}{2}\|x-v_{n,k-1}\|^2.
\end{align*}

We apply~\eqref{eq:supp-summed} at $x_n=u_n^\ast$ together with the above inequality at $x=u_n^\ast$, apply~\eqref{eq:supp-summed} at $x_n=w_{n,k-1}$ with its negative strong-convexity terms discarded, and take the convex combination of the two results with the weights $\beta$ and $1-\beta$.
Since $u_{n,k}=(1-\beta)w_{n,k-1}+\beta v_{n,k-1}$, the term involving $\Delta_{n,k}$ becomes
\[
\beta\langle\Delta_{n,k},u_n^\ast-v_{n,k-1}\rangle
 +\alpha_n^{-1}\langle\Delta_{n,k},h_n\rangle
\]
and the remaining linear term is
$\beta\langle\tilde h_n,v_{n,k-1}-v_{n,k}\rangle$.
For $n\ge1$, the remaining linear term is bounded by
$\|h_n\|^2/(2\alpha_n)+\alpha_n\beta^2\|v_{n,k}-v_{n,k-1}\|^2/2$.
For $n=0$, the first summand is replaced by
$2\|h_0\|^2/(3\alpha_0)+2\|S_k\|^2/\alpha_0$ based on
$\|h_0+S_k\|^2\le4\|h_0\|^2/3+4\|S_k\|^2$.
The squared increments $\|v_{n,k}-v_{n,k-1}\|^2$ cancel the corresponding negative quadratic terms.
For the deviation of the server, we use
\[
\beta\langle\Delta_{0,k},u_0^\ast-v_{0,k-1}\rangle
\le\frac{\beta}{2\epsilon}\|\Delta_{0,k}\|^2
 +\frac{\epsilon\beta}{2}\|u_0^\ast-v_{0,k-1}\|^2.
\]

Finally, completing the squares in $\|h_n\|$ gives
\begin{align*}
&\frac{\|S_k\|+\|\Delta_{0,k}\|}{\alpha_0}\|h_0\|
-\frac{2\alpha_0-3L_0}{6\alpha_0^2}\|h_0\|^2\\
&\quad\le\frac{3(\|S_k\|+\|\Delta_{0,k}\|)^2}{2(2\alpha_0-3L_0)}\\
&\frac{\|\Delta_{n,k}\|}{\alpha_n}\|h_n\|
-\frac{\alpha_n-L_n}{2\alpha_n^2}\|h_n\|^2\\
&\quad\le\frac{\|\Delta_{n,k}\|^2}{2(\alpha_n-L_n)},\quad n\ge1.
\end{align*}

Collecting the above steps, the RHS of~\eqref{eq:lem3} holds for each sample path with $\sigma_n^2$ replaced by $\|\Delta_{n,k}\|^2$ and with the additional term
$\sum_{n=1}^N\beta\langle\Delta_{n,k},u_n^\ast-v_{n,k-1}\rangle$.
The second moments of all iterates are finite by induction, since the smooth gradients have at most linear growth, the proximal operator is nonexpansive, the noise has bounded second moments, and the projected states are bounded.
Hence, the additional term has zero expectation by conditioning on $\mathcal F_k$ under Assumption~\ref{as:noise}.
Since neither the unbiasedness of $\Delta_{0,k}$ nor its independence of the current samples is used, taking the expectation completes the proof.

\section{Proofs of Lemmas~\ref{lem:bounded} and~\ref{lem:drift}}\label{app:stability}
\subsection{Proof of Lemma~\ref{lem:bounded}}
Let $\epsilon=\delta_0/2$ and
$\nu_{n,k}:=(\delta_n/\beta_k+\alpha_{n,k})/2$, and let $\Xi_k$ denote the sum of the last four terms on the RHS of~\eqref{eq:lem3} divided by $\beta_k^2$.
Since $D_k\ge0$ and
$(1-\beta_k)/\beta_k^2\le1/\beta_{k-1}^2$, dividing~\eqref{eq:lem3}
by $\beta_k^2$ and summing over $k=1,\ldots,K$ give
\begin{equation}\label{eq:supp-telescope}
\frac{D_K}{\beta_K^2}+\sum_{n=0}^N\nu_{n,K}\E\|u_n^\ast-v_{n,K}\|^2
\le\Lambda_0+\sum_{k=1}^K\Xi_k.
\end{equation}

Indeed, the interior coefficients of $\E\|u_n^\ast-v_{n,k}\|^2$ are
$[\alpha_{0,k+1}-\alpha_{0,k}+\epsilon/\beta_{k+1}-\delta_0/\beta_k]/2
=3\delta_0(1-k)/56$ for the server and
$[\alpha_{n,k+1}-\alpha_{n,k}-\delta_n/\beta_k]/2
=\delta_n(1-k)/28$ for the honest workers, all of which are nonpositive, and the initial terms are bounded by $\Lambda_0$ in~\eqref{eq:Lambda} since $\beta_0=1$.
Substituting the stepsizes~\eqref{eq:beta} and~\eqref{eq:alpha}, we have
$2(2\alpha_{0,k}-3L_0)\beta_k^2=8\delta_0/7$,
$2(\alpha_{n,k}-L_n)\beta_k^2=12\delta_n/7$, and
$2/(\alpha_{0,k}\beta_k^2)\le7/\delta_0$, which yield
\begin{equation}\label{eq:supp-Xi}
\begin{split}
\Xi_k\le{}&\frac{21}{8\delta_0}\E(\|S_k\|+\|\Delta_{0,k}\|)^2
 +\frac7{\delta_0}\E\|S_k\|^2\\
&+\sum_{n=1}^N\frac{7\sigma_n^2}{12\delta_n}
+\frac{k+2}{2\delta_0}\E\|\Delta_{0,k}\|^2.
\end{split}
\end{equation}

Based on~\eqref{eq:devbounds}, we have
$\|S_k\|\le Q\lambda\sqrt G$ and
$\|\Delta_{0,k}\|\le(Q+N)\lambda\sqrt G$.
With
\[
\bar\eta:=\left[7Q^2+\frac{21}8(2Q+N)^2\right]\frac{\lambda^2G}{\delta_0}
 +\sum_{n=1}^N\frac{7\sigma_n^2}{12\delta_n}
\]
we therefore obtain
\begin{align*}
&\frac{D_k}{\beta_k^2}+\sum_{n=0}^N\nu_{n,k}\E\|u_n^\ast-v_{n,k}\|^2\\
&\quad\le\Lambda_0+k\bar\eta+
\frac{(k^2+5k)(Q+N)^2\lambda^2G}{4\delta_0}.
\end{align*}

Using $4k/(k+2)^2\le1/2$ and
$(k^2+5k)/(k+2)^2\le25/24$, we have
$D_k\le\bar D:=\Lambda_0+\bar\eta/2+
25(Q+N)^2\lambda^2G/(24\delta_0)$, and the strong convexity gives
$\E\|w_{n,k}-u_n^\ast\|^2\le2\bar D/\delta_n$.
Let $\kappa_0:=\delta_0/14$ and
$\kappa_n:=3\delta_n/14$ for $n\ge1$, such that
$\nu_{n,k}\ge\kappa_n(k+2)^2/2$.
Hence, Lemma~\ref{lem:bounded} holds with
\begin{equation}\label{eq:Vn}
V_n=\max\left\{\frac{2\bar D}{\delta_n},
\frac2{\kappa_n}\left[\frac{\Lambda_0}9+\frac{\bar\eta}8+
\frac{25(Q+N)^2\lambda^2G}{96\delta_0}\right]\right\}.
\end{equation}

The same bounds hold at $k=0$, since $D_0\le\Lambda_0$ and
$\Lambda_0\ge\alpha_{n,1}\|u_n^\ast-v_{n,0}\|^2/2
\ge9\kappa_n\|u_n^\ast-v_{n,0}\|^2/2$.
Note that the above argument bounds the second moments of the iterates without assuming bounded stochastic gradients or bounded sample paths.

\subsection{Proof of Lemma~\ref{lem:drift}}
With $h_{0,k}:=\nabla f_0(u_{0,k})$ and $h_{n,k}:=\nabla f(u_{n,k};\xi_{n,k})-g_{n,k}$ for $n\ge1$, the updates imply
\[
w_{n,k}-w_{n,k-1}
=\beta_k(v_{n,k-1}-w_{n,k-1})-h_{n,k}/\alpha_{n,k}.
\]

Lemma~\ref{lem:bounded} gives
$\E\|v_{n,k-1}-w_{n,k-1}\|^2\le4V_n$ and
$\E\|u_{n,k}-u_n^\ast\|^2\le V_n$.
At the penalized optimum, we have
$\|\nabla f_0(u_0^\ast)\|\le N\lambda\sqrt G$ and
$\|\nabla f_n(u_n^\ast)\|\le\lambda\sqrt G$.
Consequently, $\E\|h_{n,k}\|^2\le\Gamma_n$ with
\begin{equation}\label{eq:E2}
\begin{split}
\Gamma_0&:=2N^2\lambda^2G+2L_0^2V_0\\
\Gamma_n&:=9\lambda^2G+6L_n^2V_n+3\sigma_n^2,\quad n\ge1
\end{split}
\end{equation}
where the bound for the honest workers uses
$\|\nabla f_n(u)+\Delta-g\|^2
\le3(\|\nabla f_n(u)\|^2+\|\Delta\|^2+\|g\|^2)$
without assuming the independence of $g$ and $\Delta$.
Using $(k+2)^2\ge9$, we obtain
\begin{align*}
\E\|w_{0,k}-w_{0,k-1}\|^2
&\le\frac{32V_0+392\Gamma_0/(9\delta_0^2)}{(k+2)^2}\\
\E\|w_{n,k}-w_{n,k-1}\|^2
&\le\frac{32V_n+392\Gamma_n/(81\delta_n^2)}{(k+2)^2}.
\end{align*}

The Lipschitz continuity of $\nabla p_n$ and the inequality
$\|a+b\|^2\le2\|a\|^2+2\|b\|^2$ yield~\eqref{eq:lem5} with
\begin{equation}\label{eq:cn}
\chi_n:=2\lambda^2L_p^2\left[32(V_0+V_n)+
\frac{392\Gamma_0}{9\delta_0^2}+
\frac{392\Gamma_n}{81\delta_n^2}\right].
\end{equation}

\section{Proof of Lemma~\ref{lem:ef} and Tightness of Its Contraction Dependence}\label{app:tracking}
\subsection{Proof of Lemma~\ref{lem:ef}}
We fix an honest worker $n$, write $\gamma:=\gamma_n$ and $q:=1-\gamma$, and omit the index $n$.
Since both $g_k$ and $\gh_{k-1}$ belong to the ball with radius $\lambda\sqrt G$, the nonexpansiveness of the projection and the contraction~\eqref{eq:general-contraction} imply the uniform cap
$\zeta_k\le4q\lambda^2G$ for $k\ge1$.
For $0<\gamma<1$, we follow the tracking recurrence of EF21~\cite[Suppl., Lemma~2]{richtarik2021}: since the projection is nonexpansive with respect to $g_k$, applying Young's inequality with the parameter $\gamma/(2q)$ to
$z_k=(g_k-g_{k-1})-(\gh_{k-1}-g_{k-1})$ and using Lemma~\ref{lem:drift}, we obtain
\[
\zeta_k\le(1-\gamma/2)\zeta_{k-1}
 +\frac{q(2-\gamma)\chi}{\gamma(k+2)^2}.
\]

For $k\le J$, the uniform cap proves~\eqref{eq:lem6}.
For $k\ge J+1$, the inductive hypothesis and $k+1\ge8/\gamma$ give
\begin{align*}
(1-\gamma/2)\left(\frac{k+2}{k+1}\right)^2
&\le(1-\gamma/2)(1+\gamma/8)^2\\
&=1-\gamma/4-7\gamma^2/64-\gamma^3/128\\
&\le1-\gamma/4
\end{align*}
and the second term of the recurrence is absorbed since
$\mathcal K\ge4q(2-\gamma)\chi/\gamma^2$.
For $\gamma=1$, the contraction makes the state of the honest worker exact per slot, which proves the result separately.
Since $J+2\le11/\gamma$, the simplified bound on $\mathcal K_n$ follows.

\subsection{Tightness of the Contraction Dependence}
The following lemma shows that the order $(1-\gamma)/\gamma^2$ in Lemma~\ref{lem:ef} cannot be improved uniformly over the bounded signals that satisfy the drift bound~\eqref{eq:lem5} and the compressors that satisfy~\eqref{eq:general-contraction}; note that it does not provide a lower bound on the trajectories generated by a specific algorithm.

\begin{lemma}[Tightness of the contraction dependence]\label{lem:tracking-sharp}
Consider the one-dimensional signal $g_k=\bar g\sin\ln(k+2)$ with $0<\bar g\le\lambda\sqrt G$, which satisfies~\eqref{eq:lem5} with $\chi=9\bar g^2/4$, and the compressor that uploads the full residual with probability $\gamma$ independently per slot.
The tracking error $\zeta_k:=\E|\gh_k-g_k|^2$ satisfies
\begin{equation}\label{eq:tracking-lower}
\limsup_{k\to\infty}\,(k+2)^2\zeta_k
=\bar g^2(1-\gamma)(2-\gamma)/\gamma^2.
\end{equation}
\end{lemma}

To prove Lemma~\ref{lem:tracking-sharp}, note that $|g_k|\le\bar g$ and
$|g_k-g_{k-1}|^2\le(9\bar g^2/4)/(k+2)^2$.
The compressor uploads $z_k$ with probability $\gamma$ and zero otherwise, which satisfies~\eqref{eq:general-contraction} with the contraction factor $\gamma$; the projection is inactive, since each state is either zero or a previous value of the signal.
With $q:=1-\gamma$, the most recent upload has the age
$i\in\{0,\ldots,k-1\}$ with probability $\gamma q^i$, and no upload has occurred with probability $q^k$.
Hence, we have
\[
\zeta_k=\gamma\sum_{i=0}^{k-1}q^i(g_{k-i}-g_k)^2+q^kg_k^2.
\]

For each fixed $i$, we have
$(k+2)(g_{k-i}-g_k)
=-\bar gi\cos\ln(k+2)+o(1)$, whose magnitude is at most $2\bar gi$ for $i\le k/2$; the contribution of the ages $i>k/2$ is bounded by a constant times
$k^2q^{k/2}$ and tends to zero.
The dominated convergence with the summable geometric weights therefore gives
\[
(k+2)^2\zeta_k
=\frac{\bar g^2q(1+q)}{\gamma^2}\cos^2\ln(k+2)+o(1).
\]

Since the cosine tends to one along the integer subsequence nearest to $\exp(2\pi j)-2$, $j=1,2,\ldots$, the limit superior is $\bar g^2(1-\gamma)(2-\gamma)/\gamma^2$ for $0<\gamma<1$, and the tracking error is zero for $\gamma=1$.

\section{Proofs of Theorems~\ref{thm:cs} and~\ref{thm:lazy}}\label{app:theorems}
\subsection{Proof of Theorem~\ref{thm:cs}}
For any $\theta>0$, the decomposition $\Delta_{0,k}=B_k+E_k$ and the bounds in~\eqref{eq:devbounds} imply
\begin{align*}
\|\Delta_{0,k}\|^2
&\le(1+\theta)B^2\lambda^2G + (1 \!+\! \frac{1}{\theta})\|E_k\|^2\\
[\|S_k\|+\|\Delta_{0,k}\|]^2
&\le(1+\theta)(Q \!+\! B)^2\lambda^2G + (1 \!+\! \frac{1}{\theta} )\|E_k\|^2.
\end{align*}

Therefore, the RHS of~\eqref{eq:supp-Xi} is at most
\begin{equation}
\eta(\theta) 
\!+\! \frac{(1 \!+\! \theta)(k \!+\! 2)B^2\lambda^2G}{2\delta_0}
\!+\! \frac{1 \!+\! 1/\theta}{8\delta_0}(4k \!+\! 29)\E\|E_k\|^2.
\end{equation}

Substituting the above bound into~\eqref{eq:supp-telescope}, discarding the nonnegative distance terms, using $\sum_{k=1}^K(k+2)=(K^2+5K)/2$, and multiplying both sides by
$\beta_K^2=4/(K+2)^2$, we obtain~\eqref{eq:thm1} with
\[
C_K=\frac{1+1/\theta}{2\delta_0(K+2)^2}
\sum_{k=1}^K(4k+29)\E\|E_k\|^2.
\]

Based on Lemma~\ref{lem:ef} and Minkowski's inequality in $L^2$, we have $\E\|E_k\|^2\le(\sum_{n=1}^N\sqrt{\mathcal K_n})^2/(k+2)^2$, which proves~\eqref{eq:CK} together with
$\sum_{k=1}^K(4k+29)/(k+2)^2
\le4\ln((K+2)/2)+21c_2$.
Fixing $\theta>0$ before taking $K\to\infty$ and then taking the infimum over $\theta$ give~\eqref{eq:limsup}.

\subsection{Proof of Theorem~\ref{thm:lazy}}
Based on~\eqref{eq:lagerror}, we have $\|E_k\|^2\le N^2\tau_k$ for each sample path.
The proof of Theorem~\ref{thm:cs} holds unchanged up to the definition of $C_K$, and the threshold~\eqref{eq:tau} gives
\[
(4k+29)\E\|E_k\|^2
\le\frac{11N^2\tau_0\lambda^2G}{\ln(k+2)}.
\]

To bound the sum over $k$, we compare it with $\int_2^{K+2}dt/\ln t$.
For $t\ge e^2$, the derivative of $2t/\ln t$ is at least $1/\ln t$; moreover, $\int_2^{e^2}dt/\ln t\le(e-2)/\ln2+e^2-e<e^2$, and the finitely many integers $K+2<e^2$ satisfy the same bound directly.
Consequently, we have $\sum_{k=1}^K1/\ln(k+2)\le2(K+2)/\ln(K+2)$, which yields~\eqref{eq:thm2}.

\section{Proof of Corollary~\ref{cor:uploads}}\label{app:uploads}
Let $\vartheta_k:=\|g_{n,k}-\gh_{n,k}\|/\omega_k$ denote the normalized tracking error of honest worker $n$.
An upload resets the tracking error to zero, and a skipped upload ensures $\vartheta_k\le1$.
Based on the trigger rule~\eqref{eq:lazy} and the triangle inequality, we have
\begin{align*}
I_{n,k}+\vartheta_k&\le\frac{\omega_{k-1}}{\omega_k}\vartheta_{k-1}+\frac{\|g_{n,k}-g_{n,k-1}\|}{\omega_k}\\
&\le\vartheta_{k-1}+\frac{\omega_{k-1}}{\omega_k}-1+\frac{\|g_{n,k}-g_{n,k-1}\|}{\omega_k}
\end{align*}
for $k\ge2$, and $I_{n,1}+\vartheta_1\le1$ regardless of the initial residual.
Summing over $k$ and dropping $\vartheta_K\ge0$ give~\eqref{eq:normalized-count} for each sample path with $\|g_{n,k}-g_{n,k-1}\|$ in place of $\sqrt{\chi_n}/(k+2)$, and taking the expectation with Jensen's inequality and Lemma~\ref{lem:drift} proves~\eqref{eq:normalized-count}.
Note that neither independence nor conditioning on stopping times is needed.

For the threshold~\eqref{eq:tau}, we have
$\omega_{k-1}/\omega_k\le(k+2)/(k+1)$ for $k\ge2$, since $\ln x/x$ decreases for $x\ge3$.
Moreover, we have $\sum_{k=2}^K(k+2)^{-1/2}\le2(\sqrt M-\sqrt3)$,
$\ln(k+2)\le\ln M$, and
$\sum_{k=2}^K1/(k+1)\le\ln((K+1)/2)$.
Substituting these bounds into~\eqref{eq:normalized-count} proves~\eqref{eq:prop1}.

For the alternative threshold discussed after Corollary~\ref{cor:uploads}, i.e.,
\begin{equation}\label{eq:nearcritical-tau}
\tau_k=\bar\tau\ln^2(k+2)/(k+2)^2
\end{equation}
we have
\begin{align}
\E[ U_{n,K} ] &\le1+\frac{2\sqrt{\chi_n/\bar\tau}M}{\ln M}+\ln\frac{K+1}{2}\label{eq:nearcritical-count}\\
C_K&\le\frac{11(1+1/\theta)N^2\bar\tau}{2\delta_0M^2}
\ln^2M\ln(M/2).\label{eq:nearcritical-cost}
\end{align}

Indeed, $\omega_k=\sqrt{\bar\tau}\ln(k+2)/(k+2)$ decreases for $k\ge1$ and obeys the same bound on $\omega_{k-1}/\omega_k$, such that
$\sum_{k=2}^K1/\ln(k+2)\le2M/\ln M$ proves~\eqref{eq:nearcritical-count}; moreover,
$(4k+29)\tau_k\le11\bar\tau\ln^2M/(k+2)$ together with $\sum_{k=1}^K1/(k+2)\le\ln(M/2)$ and the definition of $C_K$ proves~\eqref{eq:nearcritical-cost}.
Hence, the expected fraction of uploading slots vanishes as ${\cal O}(1/\ln K)$, and the additional objective error is ${\cal O}(\ln^3K/K^2)$.

\section{Implementation Details}\label{app:implementation}
\begin{table}[!t]
\centering
\caption{Final value of the common objective~\eqref{eq:penalized} at 2,000 slots in the matched-objective comparison (mean $\pm$ standard deviation over 20 seeds). Bold: lowest mean per column.}
\label{tab:matched-objective}
\setlength{\tabcolsep}{3pt}
\resizebox{\columnwidth}{!}{%
\begin{tabular}{lrrrr}
\toprule
Method & M: Clean & M: Min-Max & F: Clean & F: Min-Max\\
\midrule
Projected full & $\bestresult{6.188\pm0.132}$ & $6.770\pm0.132$ & $\bestresult{6.069\pm0.289}$ & $6.587\pm0.281$ \\
Shared reserve & $6.465\pm0.166$ & $\bestresult{6.396\pm0.167}$ & $6.393\pm0.310$ & $6.333\pm0.312$ \\
Individual top & $6.236\pm0.140$ & $6.557\pm0.133$ & $6.271\pm0.308$ & $\bestresult{6.204\pm0.306}$ \\
Uniform common & $7.279\pm0.296$ & $7.148\pm0.308$ & $7.085\pm0.366$ & $7.106\pm0.341$ \\
\bottomrule
\end{tabular}}
\end{table}

\begin{figure*}[!t]
\centering
\includegraphics[width=6.35in]{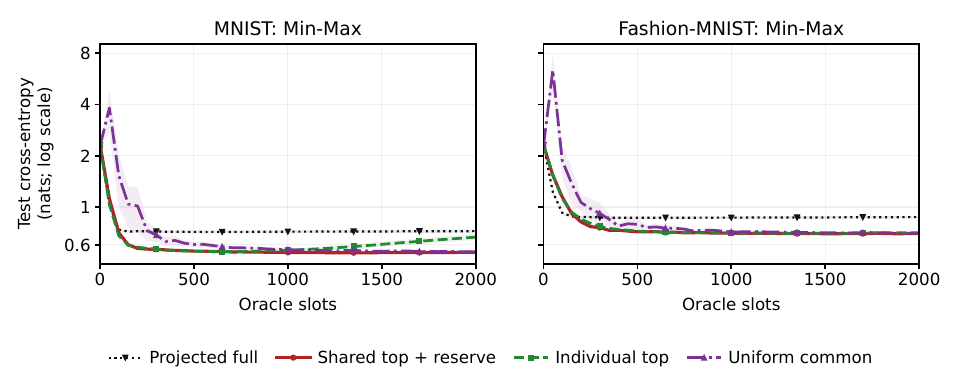}
\caption{Test cross-entropy (log scale) versus oracle slots for the four communication formats with the matched objective under the Min-Max attack. Lines and shaded regions show the means and the pointwise 95\% confidence intervals over 20 seeds.}
\label{fig:matched-convergence}
\end{figure*}
\subsubsection{\bfseries Implementation Conventions}
The proximal operator of the Huber penalty is $\prox_{tp}(v)=\mu v/(\mu+t)$ when $\|v\|\le\mu+t$, and $(1-t/\|v\|)v$ otherwise.
A valid smoothness constant of the softmax regression is $L_n=\rho+\frac1{2|\mathcal D_n|}\sum_{i\in\mathcal D_n}\|\varphi_i\|^2$ with $\varphi_i$ being the feature vector of sample $i$, since the covariance of the class probabilities has a spectral norm of at most $1/2$.
The benchmarks that aggregate loss gradients use the ridge weight $\rho+\rho_0/N=0.02$, which matches the consensus problem~\eqref{eq:consensus} normalized by $N$.
The honest workers use the exact penalty gradient in~\eqref{eq:worker-v}, including the slots with compressed or skipped uploads.
LFRPG uses a frame of $T$ local steps, the iterate of the corresponding slot in the previous frame, the current local auxiliary iterate, and the average penalty gradient of the frame; the communication occurs once per frame, and each local step is charged as an oracle slot.
FRPG and LFRPG are implemented without the projection at the server.
RSA uses the Euclidean norm penalty with the zero subgradient at coincidence and a stepsize that decays as $1/(1+k/1000)$.
The robust aggregation rules use a local exponential moving average with the momentum of zero or 0.9 selected by validation.
Krum scores the $Q-B-2$ nearest vectors of the other workers, and the trimmed mean removes $B$ entries per coordinate tail.
The geometric median uses the smoothed Weiszfeld iterations with the distance floor $10^{-6}$, the relative tolerance $10^{-7}$, and at most 30 iterations.
Byz-EF21-SGDM  \cite{liu2026momentum} initializes a full stochastic gradient estimate, which is charged on the uplink, uses the top-400 momentum residuals and the trimmed mean of the persistent states, and omits only the unused gradient evaluation after its last reported model.
FedAdam-BACK uses a Gaussian projection of rank one and eight rows, zero-fills the selected coordinates into fixed-dimensional moments, and selects the workers from the projections with the $N-1$ nearest neighbors; it applies the bias correction to the second moment only, and the offset of its denominator is $\sqrt{10^{-8}}=10^{-4}$.

\subsubsection{\bfseries Protocol and Reproduction}
The reserve rows are drawn by a pseudorandom generator, which does not provide a cryptographic guarantee of unpredictability.
The split seed is 20261103, the calibration seed is 9001, the seeds of the exploratory study are 0--4, and the seeds of the main comparison are 100--119.
The candidates of each benchmark vary the penalty parameters, the learning rate, the momentum, the compression budget, or the local frame as appropriate; hence, the tuning provides the same number of candidates rather than equally expressive search spaces.
A budget endpoint is the last model before the next communication would exceed the cap on the uploaded bytes averaged over the honest workers, under the common ceiling of 2,000 slots; a lazy worker can individually exceed the average.
The traffic counts the numerical content and the indices, where the controlled study uses 8-bit row indices; headers, retransmissions, and latency are not modeled.
The supplementary archive contains the code with pinned dependencies and dataset checksums, the candidate grids, the final models and the trajectories of all runs, and the scripts that recompute the reported metrics and the paired statistics.

\subsubsection{\bfseries Matched-Objective Comparison}
All four formats in Section~\ref{sec:matched} use the identical $\lambda=0.5$, $\mu=0.1$, stepsizes, partitions, and mini-batch streams.
The three sparse formats transmit 16 rows of length 50, where the shared mask uses 13 ranked rows and three reserve rows with $\widehat b=B$, the individual top rows upload their own indices, and the uniformly sampled common rows require no energy reports.
Each format enforces a different feasible support under the same Min-Max rule, and the shared mask additionally admits forged energy reports.
The 80 runs with the shared mask reproduce the corresponding runs of the main comparison exactly.
Table~\ref{tab:matched-objective} reports the final value of the common objective~\eqref{eq:penalized}, and Fig.~\ref{fig:matched-convergence} shows the test cross-entropy of the four formats under the Min-Max attack.

\subsubsection{\bfseries Controlled Study}
The minimizer of~\eqref{eq:penalized} is approximated by an L-BFGS solution $w_{\rm ref}$, which satisfies the certificate
\begin{equation}\label{eq:test-reference}
0\le F(w_{\rm ref})-F(u^\ast)
\le\|\nabla F(w_{\rm ref})\|^2/(2\delta_{\min})
\end{equation}
with the RHS below $3.43\times10^{-9}$ in all cases.
With eight rows under two sign-flipping Byzantine workers, the individual top rows upload 5.10\% of the dense vector plus 64 index bits per slot, compared with 7.10\% for the shared mask, and attain the accuracy of $88.35\pm0.64$\% versus $87.98\pm0.90$\% on MNIST and $78.60\pm1.51$\% versus $78.63\pm1.71$\% on Fashion-MNIST over five seeds.
The 20,000-slot runs of LAG-FRPG show a decreasing upload activity, although they are too short to estimate the asymptotic exponent in Corollary~\ref{cor:uploads}.

\subsubsection{\bfseries Consensus Diagnostics}
Let $\bar u$ and $\bar x$ denote the numerical solutions of the penalized problem~\eqref{eq:penalized} and the consensus problem~\eqref{eq:consensus}, respectively, with
$\varepsilon_F:=\|\nabla F(\bar u)\|/\delta_{\min}$ and
$\varepsilon_H:=\|\nabla H(\bar x)\|/\delta_\Sigma$.
The exact model bias $\|u_0^\ast-x^\star\|$ lies between
$[\|\bar u_0-\bar x\|-\varepsilon_F-\varepsilon_H]_+$ and
$\|\bar u_0-\bar x\|+\varepsilon_F+\varepsilon_H$, which gives the enclosures in Table~\ref{tab:consensus-diagnostics}.
Over the 20 reference problems, the largest gradient norm of the consensus solutions is $7.903\times10^{-7}$, and the largest certificate~\eqref{eq:consensus-residual-objective} of the consensus objective is $1.562\times10^{-12}$.

\ifdefined\TSPsplit\else
  \printreferences
\fi


\begin{thebibliography}{10}
\providecommand{\url}[1]{#1}
\csname url@samestyle\endcsname
\providecommand{\newblock}{\relax}
\providecommand{\bibinfo}[2]{#2}
\providecommand{\BIBentrySTDinterwordspacing}{\spaceskip=0pt\relax}
\providecommand{\BIBentryALTinterwordstretchfactor}{4}
\providecommand{\BIBentryALTinterwordspacing}{\spaceskip=\fontdimen2\font plus
\BIBentryALTinterwordstretchfactor\fontdimen3\font minus
  \fontdimen4\font\relax}
\providecommand{\BIBforeignlanguage}[2]{{%
\expandafter\ifx\csname l@#1\endcsname\relax
\typeout{** WARNING: IEEEtran.bst: No hyphenation pattern has been}%
\typeout{** loaded for the language `#1'. Using the pattern for}%
\typeout{** the default language instead.}%
\else
\language=\csname l@#1\endcsname
\fi
#2}}
\providecommand{\BIBdecl}{\relax}
\BIBdecl

\bibitem{mcmahan2017}
B.~McMahan, E.~Moore, D.~Ramage, S.~Hampson, and B.~{Ag\"uera y Arcas},
  ``{Communication-efficient learning of deep networks from decentralized
  data},'' in \emph{Proc. AISTATS}, ser. PMLR, vol.~54, 2017, pp. 1273--1282.

\bibitem{blanchard2017}
P.~Blanchard, E.~M. El~Mhamdi, R.~Guerraoui, and J.~Stainer, ``{Machine
  learning with adversaries: Byzantine tolerant gradient descent},'' in
  \emph{Adv. Neural Inf. Process. Syst.}, vol.~30, 2017.

\bibitem{yin2018}
D.~Yin, Y.~Chen, K.~Ramchandran, and P.~Bartlett, ``{Byzantine-robust
  distributed learning: Towards optimal statistical rates},'' in \emph{Proc.
  ICML}, ser. PMLR, vol.~80, 2018, pp. 5650--5659.

\bibitem{stich2018}
S.~U. Stich, J.-B. Cordonnier, and M.~Jaggi, ``{Sparsified SGD with memory},''
  in \emph{Adv. Neural Inf. Process. Syst.}, vol.~31, 2018.

\bibitem{chen2018lag}
T.~Chen, G.~B. Giannakis, T.~Sun, and W.~Yin, ``{LAG: Lazily aggregated
  gradient for communication-efficient distributed learning},'' in \emph{Adv.
  Neural Inf. Process. Syst.}, vol.~31, 2018.

\bibitem{mhamdi2018bulyan}
E.~M. El~Mhamdi, R.~Guerraoui, and S.~Rouault, ``{The hidden vulnerability of
  distributed learning in Byzantium},'' in \emph{Proc. ICML}, ser. PMLR,
  vol.~80, 2018, pp. 3521--3530.

\bibitem{pillutla2022rfa}
K.~Pillutla, S.~M. Kakade, and Z.~Harchaoui, ``{Robust aggregation for
  federated learning},'' \emph{IEEE Trans. Signal Process.}, vol.~70, pp.
  1142--1154, 2022.

\bibitem{wu2020byrdsaga}
Z.~Wu, Q.~Ling, T.~Chen, and G.~B. Giannakis, ``{Federated variance-reduced stochastic gradient descent with robustness to Byzantine attacks},''
  \emph{IEEE Trans. Signal Process.}, vol.~68, pp. 4583--4596, 2020.

\bibitem{li2019rsa}
L.~Li, W.~Xu, T.~Chen, G.~B. Giannakis, and Q.~Ling, ``{RSA: Byzantine-robust
  stochastic aggregation methods for distributed learning from heterogeneous
  datasets},'' in \emph{Proc. AAAI Conf. Artificial Intelligence}, vol.~33,
  no.~1, 2019, pp. 1544--1551.

\bibitem{dong2020frpg}
Y.~Dong, G.~B. Giannakis, T.~Chen, J.~Cheng, M.~J. Hossain, and V.~C.~M. Leung,
  ``{Communication-efficient Byzantine-resilient federated learning over
  heterogeneous datasets},'' arXiv:2006.09992v3, 2020.

\bibitem{karimireddy2019ef}
S.~P. Karimireddy, Q.~Rebjock, S.~U. Stich, and M.~Jaggi, ``{Error feedback fixes SignSGD and other gradient compression schemes},'' in \emph{Proc.
  ICML}, ser. PMLR, vol.~97, 2019, pp. 3252--3261.

\bibitem{richtarik2021}
P.~Richt\'{a}rik, I.~Sokolov, and I.~Fatkhullin, ``{EF21: A new, simpler,
  theoretically better, and practically faster error feedback},'' in \emph{Adv.
  Neural Inf. Process. Syst.}, vol.~34, 2021.

\bibitem{yang2018gboard}
T.~Yang \emph{et~al.}, ``{Applied federated learning: Improving Google keyboard
  query suggestions},'' arXiv:1812.02903, 2018.

\bibitem{legate2023lastlayer}
G.~Legate, N.~Bernier, L.~Page-Caccia, E.~Oyallon, and E.~Belilovsky,
  ``{Guiding the last layer in federated learning with pre-trained models},''
  in \emph{Adv. Neural Inf. Process. Syst.}, vol.~36, 2023.

\bibitem{chen2019sage}
Y.~Chen, S.~Kar, and J.~M.~F. Moura, ``{Resilient distributed parameter
  estimation with heterogeneous data},'' \emph{IEEE Trans. Signal Process.},
  vol.~67, no.~19, pp. 4918--4933, Oct. 2019.

\bibitem{liu2011fdi}
Y.~Liu, P.~Ning, and M.~K. Reiter, ``{False data injection attacks against
  state estimation in electric power grids},'' \emph{ACM Trans. Inf. Syst.
  Secur.}, vol.~14, no.~1, May 2011, Art. no. 13.

\bibitem{karimireddy2021history}
S.~P. Karimireddy, L.~He, and M.~Jaggi, ``{Learning from history for Byzantine robust optimization},'' in \emph{Proc. ICML}, vol. 139, 2021, pp. 5311--5319.

\bibitem{karimireddy2022bucketing}
------, ``{Byzantine-robust learning on heterogeneous datasets via bucketing},'' in \emph{Proc. ICLR}, 2022.

\bibitem{allouah2023mixing}
Y.~Allouah, S.~Farhadkhani, R.~Guerraoui, N.~Gupta, R.~Pinot, and J.~Stephan,
  ``{Fixing by mixing: A recipe for optimal Byzantine ML under heterogeneity},'' in \emph{Proc. AISTATS}, vol. 206, 2023, pp. 1232--1300.

\bibitem{allouah2023breakdown}
Y.~Allouah, R.~Guerraoui, N.~Gupta, R.~Pinot, and G.~Rizk, ``{Robust distributed learning: Tight error bounds and breakdown point under data heterogeneity},'' in \emph{Adv. Neural Inf. Process. Syst.}, vol.~36, 2023,
  pp. 45\,744--45\,776.

\bibitem{shi2025optimal}
Q.~Shi, J.~Peng, K.~Yuan, X.~Wang, and Q.~Ling, ``{Optimal complexity in Byzantine-robust distributed stochastic optimization with data heterogeneity},'' \emph{J. Mach. Learn. Res.}, vol.~26, no. 268, pp. 1--58,
  2025.

\bibitem{cao2021fltrust}
X.~Cao, M.~Fang, J.~Liu, and N.~Z. Gong, ``{FLTrust: Byzantine-robust federated
  learning via trust bootstrapping},'' in \emph{Proc. NDSS}, 2021.

\bibitem{xie2020zenopp}
C.~Xie, S.~Koyejo, and I.~Gupta, ``{Zeno++: Robust fully asynchronous SGD},''
  in \emph{Proc. ICML}, ser. PMLR, vol. 119, 2020, pp. 10\,495--10\,503.

\bibitem{shestakov2026jl}
A.~Shestakov, V.~Aletov, and A.~Beznosikov, ``{On Johnson-Lindenstrauss
  transforms for Byzantine-robust optimization},'' arXiv:2610.27989v1, 2026.

\bibitem{xie2019ipm}
C.~Xie, O.~Koyejo, and I.~Gupta, ``{Fall of empires: Breaking
  Byzantine-tolerant SGD by inner product manipulation},'' in \emph{Proc. 35th
  Conf. Uncertainty in Artificial Intelligence}, ser. PMLR, vol. 115, 2020, pp.
  261--270.

\bibitem{baruch2019alie}
M.~Baruch, G.~Baruch, and Y.~Goldberg, ``{A little is enough: Circumventing
  defenses for distributed learning},'' in \emph{Adv. Neural Inf. Process.
  Syst.}, vol.~32, 2019.

\bibitem{shejwalkar2021minmax}
V.~Shejwalkar and A.~Houmansadr, ``{Manipulating the Byzantine: Optimizing
  model poisoning attacks and defenses for federated learning},'' in
  \emph{Proc. NDSS}, 2021.

\bibitem{koppel2017}
A.~Koppel, B.~M. Sadler, and A.~Ribeiro, ``{Proximity without consensus in
  online multiagent optimization},'' \emph{IEEE Trans. Signal Process.},
  vol.~65, no.~12, pp. 3062--3077, 2017.

\bibitem{dinh2020pfedme}
C.~T. Dinh, N.~Tran, and J.~Nguyen, ``{Personalized federated learning with
  Moreau envelopes},'' in \emph{Adv. Neural Inf. Process. Syst.}, vol.~33,
  2020, pp. 21\,394--21\,405.

\bibitem{ghosh2021ef}
A.~Ghosh, R.~K. Maity, S.~Kadhe, A.~Mazumdar, and K.~Ramchandran,
  ``{Communication-efficient and Byzantine-robust distributed learning with error feedback},'' \emph{IEEE J. Sel. Areas Inf. Theory}, vol.~2, no.~3, pp.
  942--953, 2021.

\bibitem{zhu2023broadcast}
H.~Zhu and Q.~Ling, ``{Byzantine-robust distributed learning with compression},'' \emph{IEEE Trans. Signal Inf. Process. Netw.}, vol.~9, pp.
  280--294, 2023.

\bibitem{rammal2024}
A.~Rammal, K.~Gruntkowska, N.~Fedin, E.~Gorbunov, and P.~Richt\'{a}rik,
  ``{Communication compression for Byzantine robust learning: New efficient
  algorithms and improved rates},'' in \emph{Proc. AISTATS}, ser. PMLR, vol.
  238, 2024, pp. 1207--1215.

\bibitem{liu2026momentum}
C.~Liu, Y.~Li, Y.~Yi, and K.~H. Johansson, ``{Byzantine-robust and communication-efficient distributed learning via compressed momentum filtering},'' \emph{IEEE Trans. Neural Netw. Learn. Syst.}, vol.~37, no.~8,
  pp. 3985--3997, 2026.

\bibitem{li2026byzdm21}
Y.~Li, C.~Liu, and Y.~Yi, ``{Accelerating Byzantine-robust distributed learning with compressed communication via double momentum and variance reduction},''
  arXiv:2603.15144v2, 2026.

\bibitem{gupta2026rosdhb}
D.~Gupta, A.~Honsell, C.~Xu, N.~Gupta, and G.~Neglia, ``{Reconciling communication compression and Byzantine-robustness in distributed learning},'' in \emph{Proc. AISTATS}, vol. 300, 2026, pp. 2008--2016.

\bibitem{xu2026fedadamback}
L.~Xu, J.~Chen, H.~Ko, W.~Wang, and Y.~Dong, ``{FedAdam-BACK: A
  compression-based Byzantine-resilient federated learning},'' 2026,
  unpublished manuscript.

\bibitem{chen2025arctopk}
C.~Chen, C.~Ma, Z.~Li, Y.~He, Y.~Dong, and K.~Yuan, ``{An All-Reduce compatible
  Top-$K$ compressor for communication-efficient distributed learning},''
  arXiv:2510.26709v3, 2025.

\bibitem{he2023crsa}
X.~He, H.~Zhu, and Q.~Ling, ``{C-RSA: Byzantine-robust and
  communication-efficient distributed learning in the non-convex and non-IID
  regime},'' \emph{Signal Process.}, vol. 213, Dec. 2023, art. no. 109222.

\bibitem{chen2021lasg}
T.~Chen, Y.~Sun, and W.~Yin, ``{Communication-adaptive stochastic gradient
  methods for distributed learning},'' \emph{IEEE Trans. Signal Process.},
  vol.~69, pp. 4637--4651, 2021.

\bibitem{sun2022laq}
J.~Sun, T.~Chen, G.~B. Giannakis, Q.~Yang, and Z.~Yang, ``{Lazily aggregated
  quantized gradient innovation for communication-efficient federated
  learning},'' \emph{IEEE Trans. Pattern Anal. Mach. Intell.}, vol.~44, no.~4,
  pp. 2031--2044, Apr. 2022.

\bibitem{singh2023sparq}
N.~Singh, D.~Data, J.~George, and S.~Diggavi, ``{SPARQ-SGD: Event-triggered and
  compressed communication in decentralized optimization},'' \emph{IEEE Trans.
  Autom. Control}, vol.~68, no.~2, pp. 721--736, 2023.

\bibitem{he2023event}
X.~He, X.~Yi, Y.~Zhao, K.~H. Johansson, and V.~Gupta, ``{Asymptotic analysis of
  federated learning under event-triggered communication},'' \emph{IEEE Trans.
  Signal Process.}, vol.~71, pp. 2654--2667, 2023.

\bibitem{mishchenko2022proxskip}
K.~Mishchenko, G.~Malinovsky, S.~Stich, and P.~Richt\'{a}rik, ``{ProxSkip: Yes! Local gradient steps provably lead to communication acceleration! Finally!}''
  in \emph{Proc. ICML}, ser. PMLR, vol. 162, 2022, pp. 15\,750--15\,769.

\bibitem{condat2025locodl}
L.~Condat, A.~Maranjyan, and P.~Richt\'{a}rik, ``{LoCoDL: Communication-efficient distributed learning with local training and compression},'' in \emph{Proc. ICLR}, 2025.

\bibitem{mnist}
Y.~LeCun, L.~Bottou, Y.~Bengio, and P.~Haffner, ``{Gradient-based learning
  applied to document recognition},'' \emph{Proc. IEEE}, vol.~86, no.~11, pp.
  2278--2324, 1998.

\bibitem{xiao2017fashion}
H.~Xiao, K.~Rasul, and R.~Vollgraf, ``{Fashion-MNIST: A novel image dataset for
  benchmarking machine learning algorithms},'' arXiv:1708.07747, 2017.

\end{thebibliography}
\end{document}